%% file: surface_cnn.tex
\documentclass[10pt,twocolumn,letterpaper]{article}

\usepackage{cvpr}
\usepackage{times}
\usepackage{epsfig}
\usepackage{graphicx}
\usepackage{amsmath}
\usepackage{amssymb}
\usepackage{overpic}
\usepackage[ruled]{algorithm2e}
\usepackage{authblk}

\graphicspath{{./images/}}

\usepackage[pagebackref=true,breaklinks=true,letterpaper=true,colorlinks,bookmarks=false]{hyperref}

\cvprfinalcopy 

\makeatletter \renewcommand\AB@affilsepx{\qquad \protect\Affilfont} \makeatother

\renewcommand*{\Affilfont}{\normalsize}

\ifcvprfinal\fi
\begin{document}

\title{PFCNN: Convolutional Neural Networks on 3D Surfaces Using Parallel Frames}

\author[1,3]{\vspace{-7mm}Yuqi Yang\thanks{Joint first author. Work done during internship at Microsoft.}}
\author[2,3]{Shilin Liu\protect\footnotemark[1]}
\author[3]{Hao Pan\thanks{Corresponding author.}}
\author[3]{Yang Liu}
\author[3]{Xin Tong\vspace{-4mm}}
\affil[1]{Tsinghua University}
\affil[2]{University of Science and Technology of China}
\affil[3]{Microsoft Research Asia}
\affil[ ]{\tt\small yangyq18@mails.tsinghua.edu.cn freelin@mail.ustc.edu.cn \{haopan,yangliu,xtong\}@microsoft.com\vspace{-7mm}}

\maketitle

\begin{abstract}
   Surface meshes are widely used shape representations and capture finer geometry data than point clouds or volumetric grids, but are challenging to apply CNNs directly due to their non-Euclidean structure.
   We use parallel frames on surface to define PFCNNs that enable effective feature learning on surface meshes by mimicking standard convolutions faithfully.
   In particular, the convolution of PFCNN not only maps local surface patches onto flat tangent planes, but also aligns the tangent planes such that they locally form a flat Euclidean structure, thus enabling recovery of standard convolutions.
   The alignment is achieved by the tool of locally flat connections borrowed from discrete differential geometry, which can be efficiently encoded and computed by parallel frame fields.
   In addition, the lack of canonical axis on surface is handled by sampling with the frame directions.
   Experiments show that for tasks including classification, segmentation and registration on deformable geometric domains, as well as semantic scene segmentation on rigid domains, PFCNNs achieve robust and superior performances without using sophisticated input features than state-of-the-art surface based CNNs.
\end{abstract}

\input{src/introduction}

\input{src/relatedwork}

\input{src/method}

\input{src/results}

\input{src/conclusion}


{\small
\bibliographystyle{ieee_fullname}
\bibliography{surface_cnn}
}

\newpage
\clearpage

\input{src/appendix}

\end{document}

%% file: src/introduction.tex
\section{Introduction}

Applying CNNs to 3D geometric domains is critical for deep learning beyond the 2D images.
Unlike regular 2D images, 3D geometric data can be represented in different forms, posing challenges to standard CNNs.
For example, volumetric grids regularly sample $\mathbf{R}^3$ on which CNNs can be trivially deployed, but they are memory consuming and inflexible for capturing fine geometric details. 
For representation efficiency, 3D objects and scenes are frequently encoded by their boundary surfaces discretized as triangle meshes. 
However, the curved and irregularly sampled meshes do not admit the standard CNNs designed for flat image domains with regular pixel grids.
While several surface based CNNs have been proposed to tackle this problem, in this paper we use parallel frame fields that contain pointwise $N$-direction frames (Fig.~\ref{fig:parallel_frames}) to define a novel PFCNN framework whose convolution mimics standard image convolutions more faithfully.

Similar to standard CNNs, the PFCNN convolution works on a local surface patch each time and maps it onto the flat tangent space where the convolution kernel is parameterized, as done by many previous surface based CNNs \cite{GeoCNN:ICCV2015,AnisoCNN:NIPS2016,MoNet2017,Poulenard:2018:Multidirectional}.
Different from the previous approaches, however, we also align the tangent spaces of different surface points such that they locally form a flat Euclidean structure, on which the surface-based feature maps and convolution kernels can be moved as in the standard image domain.
For images, such translation operations are formally captured by the translation equivariance property of convolution \cite{GeometricDL2017}, which is a key factor contributing to the effectiveness of CNNs by enabling shared trainable weights and thus significantly reducing the amount of network parameters to avoid overfitting and achieve generalization \cite{LeCun98,Goodfellow-et-al-2016};
our surface based convolution is shown to reproduce the image-domain translation equivariance locally.

We adopt the tool of locally flat connections from discrete differential geometry \cite{DirFieldEG2016} to align the tangent spaces.
The locally flat connection is encoded by the field of pointwise tangential $N$-direction frames  (Fig.~\ref{fig:parallel_frames}) that is efficiently computed to be parallel and aligned to salient geometric features to better capture semantics.
In addition, because there exists no canonical axis on a surface, we sample the axes using the same $N$ frame directions and organize the resulting feature maps with an $N$-cover space of the domain surface \cite{dubrovin2012modern}; on each sheet of the cover space, the canonical axis is selected and the convolution is readily defined.
Furthermore, to handle the irregular mesh vertices, for each patch we resample with a regular grid and apply standard shaped convolution kernels on it.

The PFCNNs resemble standard CNNs so that efficient network structures can be leveraged accordingly.
Through experiments of deformable shape classification, segmentation and matching as well as rigid scene segmentation, we show that PFCNNs using only raw input signals achieve superior performances than competing surface CNN frameworks.
In addition, we do extensive ablation studies to validate the components of our framework.

%% file: src/relatedwork.tex
\section{Related work}
\label{sec:relatedworks}

We briefly review 3D neural networks by classifying them according to the forms of domain representation, and focus on the most related works that use surface meshes.

\vspace{-3.5mm}
\paragraph{3D neural networks for volumetric grids, point clouds and multi-view representations.}
The earliest works for 3D deep learning directly extend CNNs to 3D volumetric grids \cite{wu20153d,maturana2015voxnet}, which are later improved for computational efficiency by using adaptive grids like octrees that use high resolution only around the boundary surfaces \cite{riegler2017octnet,wang2017ocnn}. 
Point sets also conveniently encode 3D shapes, for which the set-based PointNet \cite{qi2017pointnet} is proposed and later extended by PointNet++ \cite{qi2017pointnetplusplus} to take advantage of the local surface patch structure.
Similarly, more works utilize the local patch structures of 3D point clouds, by e.g. tangent plane projection \cite{Koltun:2018:TangentConv}, localization with lattice structure \cite{su2018splatnet}, or localized kernel functions \cite{Atzmon:2018:PCN,thomas2019KPConv}.
Multi-view representations encode 3D data with a set of 2D images \cite{su15mvcnn,qi2016volumetric}, on which standard CNNs are applied to extract intermediate features and aggregated for final output.
The PFCNNs presented in this paper work on surface meshes, which have been used pervasively for 3D representation due to their high efficiency for capturing geometry to the fine details.

\vspace{-3.5mm}
\paragraph{Patch-based surface CNNs.}
A series of works extend standard CNNs to curved surface domains by applying convolution operations on localized geodesic patches; they differ mainly in the specific ways of convolution computation.
Masci et al.~\cite{GeoCNN:ICCV2015} parameterize each geodesic patch in polar coordinates, upon which the convolution operation is computed by rotating the kernel function for a set of discrete angles and convolving it with input features; the convolved features for different angles are further pooled for output.
With such an approach, it is hard to capture anisotropic or directional signals.
Later Boscaini et al.~\cite{AnisoCNN:NIPS2016} propose anisotropic CNNs that extend \cite{GeoCNN:ICCV2015} by aligning the convolution kernels to frames of principal curvature directions, thus removing angular pooling and ambiguity, and  show improved performances on various tasks.
Xu et al. \cite{Zhong:DirectionalCNN:2017} use a similar convolution on $n$-ring neighboring faces with fixed cardinality for shape segmentation.
MoNet \cite{MoNet2017} extends the geodesic convolutions by modeling the convolution kernel as a mixture of Gaussians whose bases and coefficients are fully trainable rather than functions of fixed parameterizations.
TextureNet~\cite{Huang_2019_CVPR} imposes locally rectangular grids define by 4-directional fields on the geodesic patches, and extract the features for center or corner grid points separately to handle the grid orientation ambiguity.
Multi-directional CNNs \cite{Poulenard:2018:Multidirectional} make the further step of resolving the orientation alignment of geodesic patches by using parallel transport to match the directional convolution responses for different surface points, which enables effective propagation of directional signals.
Different from these manifold based works, SplineCNN \cite{SplineCNN} defines 3D spline convolution kernels for extracting features on surface and is inherently a volumetric approach focusing on handling the irregular sampling of meshes.

Our PFCNN follows the geodesic convolution paradigm, but differs from others in the convolution computation.
Indeed, our framework closely relates to the latest parallel transport approach of \cite{Poulenard:2018:Multidirectional}, but we align the tangent spaces with locally flat connections that not only approximate the parallel transport but also induce a locally Euclidean structure suitable for defining convolutions as for images.
In addition, the locally flat connections can be adapted to capture salient geometric features like sharp creases, which further improves performance.
As a result, our PFCNNs show superior performances than the previous patch-based surface CNNs on diverse tasks (Sec.~\ref{sec:results}).

\vspace{-3.5mm}
\paragraph{Surface CNNs using atlas maps.}
Another series of works deal with a surface domain by mapping it to a 2D atlas image, on which standard convolutions are applied.
Sinha et al.~\cite{GeometryImage2016} use the geometry image to map a 3D surface of sphere topology to the planar domain and feed the map to CNNs for shape recognition.
Maron et al.~\cite{Maron:2017:ToricCoverCNN} note the geometry images have gaps between charts of the atlas map and propose to parameterize a surface of sphere topology conformally to the flat image with a toric topology, where standard convolutions with cross boundary cyclic padding are applied. 
Such convolutions are shown to be conformally translation equivariant but the conformal scaling distortion is uneven for different surface regions.
Li et al.~\cite{li2019cross} handle the gaps of an atlas map by modulating the convolution to jump across the gaps, while the mapping distortion is loosely constrained by subdividing the charts.
In comparison, our framework works with surfaces of general topology and automatically preserves the original signals with minimal distortion due to the local patch paradigm.

%% file: src/method.tex
\section{Overview}

\begin{figure}
	\centering
	\begin{overpic}[width=0.85\linewidth]{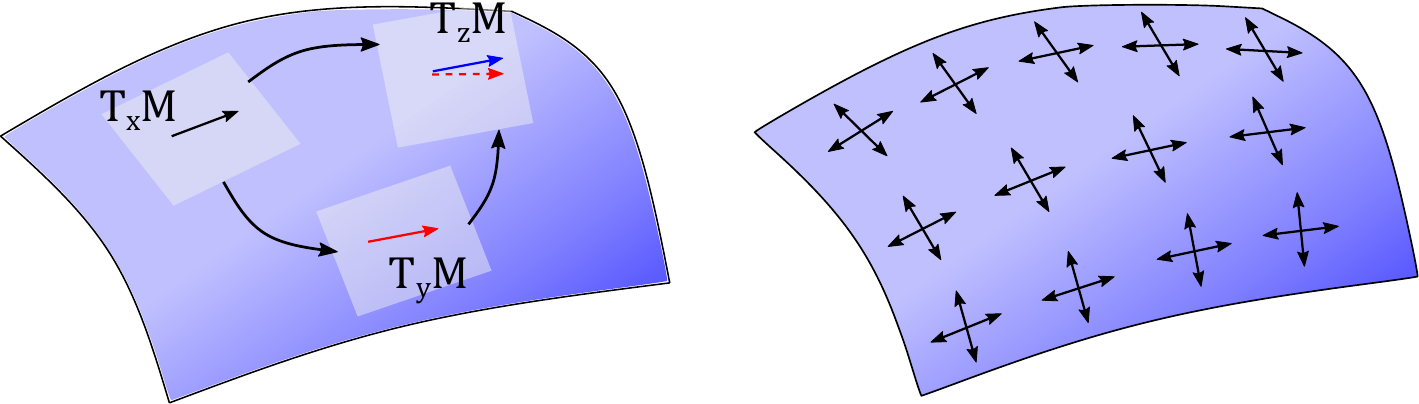}
		\put(2,1){\small (a)}
		\put(55,1){\small (b)}
	\end{overpic}
	\vspace{0mm}
	\caption{For patch-based surface CNNs, the key problem is how to align the tangent spaces of different surface points. (a) the parallel transport is path-dependent and maps the vector in $T_x\mathcal{M}$ directly to the blue one in $T_z\mathcal{M}$ but to the red dashed one by going through $T_y\mathcal{M}$. (b) by building a flat connection encoded by the parallel 4-direction frame field, our approach has path-independent translation as in image domain.}
	\label{fig:parallel_frames}
	\vspace{-5mm}
\end{figure}

To represent the boundary of a 3D object, we consider a surface mesh $\mathcal{M} = (V,F)$, with $V=\{v_i\}$ the vertices with embedding $\mathbf{v}_i\in\mathbf{R}^3$, and $F=\{f_i=(v_{i0},v_{i1},v_{i2})\}$ the faces with corners indexing the vertices.
Denote the unit normal vector at vertex $v_i$ as $\mathbf{n}_{v_i}\in\mathbf{R}^3$, and tangent plane as $T_{v_i}\mathcal{M}$ on which we can project the local geodesic patch and apply standard image-like convolutions.
As reviewed in Sec.~\ref{sec:relatedworks}, while most patch-based surface CNNs follow this general approach, the key challenge is how to coordinate the convolutions for tangent planes of different vertices (Fig.~\ref{fig:parallel_frames}).
We resolve this challenge by building locally flat connections that align the tangent planes into locally flat Euclidean domains, thus enabling effective weight sharing and translation equivariance that mimic behavior on 2D images.

In Sec.~\ref{sec:background} we briefly review the standard Euclidean convolutions with their translation equivariance property, the notion of connections from differential geometry, locally flat connections encoded by $N$-direction frame fields and $N$-cover spaces for organizing convolution and feature maps.
In Sec.~\ref{sec:model} we present the extended convolution on surfaces using parallel frames that achieves local translation equivariance and handles irregular vertex sampling on meshes, and the new layers that constitute a PFCNN model.

\section{Background}
\label{sec:background}

\subsection{Convolution on Euclidean domains}
The convolution operation of a CNN exploits the translation equivariance of 2D images \cite{LeCun89,LeCun98,GeometricDL2017}.
Let $f,k: \Omega\subset\mathbf{R}^2\rightarrow \mathbf{R}$ be two functions defined on the image $\Omega$, and $k$ is the convolution kernel usually with a local spatial support.
Define the convolution operator $\star$ as $f\star k(x)=\int_{y\in\Omega}{k(y-x)f(y)dy}$.
A planar translation of the image-based function by a vector $v\in\mathbf{R}^2$ is $\tau_v(f(x))=f(x-v)$.
Translation equivariance simply means that the planar translation commutes with convolution, i.e.
\begin{equation}
\label{eqn:euclidean_trans}
\tau_v(f\star k) = \tau_v(f)\star k.
\end{equation}
CNNs parameterize the convolution kernel with trainable weights, which can be shared for different image regions and therefore lead to less overfitting and more generality.
As will be discussed next, on curved surface domains the notion of translation is only locally meaningful, which poses difficulty for effective weight sharing of the convolution kernels.

\subsection{Connections and locally flat connections}
Connections generalize the notion of translation onto curved manifolds with non-Euclidean metric \cite{Lee1997}.
Intuitively, a (linear) connection $\nabla: T\mathcal{M}\times T\mathcal{M} \rightarrow T\mathcal{M}$ measures the linear differentiation of moving tangent plane $T_x\mathcal{M}$ along a vector $v \in T_x\mathcal{M}$ infinitesimally.
Therefore, a geodesic curve $\gamma: [0,1]\rightarrow \mathcal{M}$ as the ``straight line'' on a surface has $\nabla_{\dot{\gamma}}\dot{\gamma} = 0$, i.e. the curve tangent vector moves straightly along itself.
Indeed, the patch-based multi-directional geodesic CNN(MDGCNN) \cite{Poulenard:2018:Multidirectional} connects the convolutions for two surface patches by translating the tangent planes along the geodesic curve connecting the two patch centers, which provides a natural extension of translation on 2D images.

However, the problem with parallel transporting along the geodesic curves is that the mapping is path dependent.
Consider three nearby points $x,y,z\in\mathcal{M}$, and denote the transport of tangent planes along the geodesic curve between $x,y$ as $\tau_{x,y}: T_x\mathcal{M} \rightarrow T_y\mathcal{M}$.
In general, we have $\tau_{y,z}\circ\tau_{x,y} \neq \tau_{x,z}$, where $\circ$ is composition, with the difference caused by the curvature of the triangular surface patch bounded by the geodesic curves (Fig.~\ref{fig:parallel_frames}(a)).

In this paper, we propose to use a construction called locally flat (or trivial) connections \cite{Crane:TrivialConnection,Ray:GeometryAware} to achieve the path-independent tangent space mapping for all surface patches except at a few singular points.
The idea of locally flat connections is to concentrate the surface curvature onto a sparse set of singular points and leave the rest majority of surface area with tangent space mappings as in a Euclidean domain, which in turn paves the way for convolutions as on images.

\subsection{$N$-direction frame fields and cover space}
\label{sec:frame_field}
One way of encoding the locally flat connections for meshes is through $N$-direction frame fields \cite{Ray:GeometryAware,DirFieldEG2016}.
An $N$-direction field at $x\in\mathcal{M}$ gives a frame of $N$ rotationally symmetric directions $\mathbf{u}_x^i\in T_x\mathcal{M}, i=1,\cdots,N$; thus two consecutive vectors in the sequence differ by an angle $\frac{2\pi}{N}$.
A transport (or matching) $\tau_{x,y}$ between two tangent planes of $x,y$ can thus be defined by identifying $\mathbf{u}_x^i$ with $\mathbf{u}_y^j$, which simply amounts to a change of bases.
In particular, we use the \textit{principal matching} which chooses $j$ such that $\|\tau'_{x,y}(\mathbf{u}_x^i) - \mathbf{u}_y^j\|$ is minimal, where $\tau'_{x,y}$ is the parallel transport between $x,y$ along geodesics.

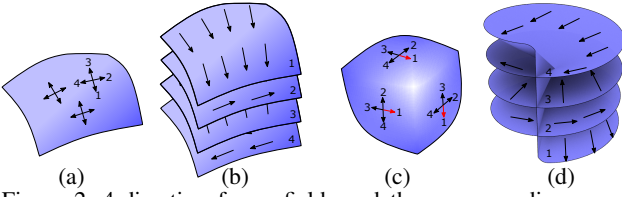
\begin{figure}
	\centering
	\begin{overpic}[width=\linewidth]{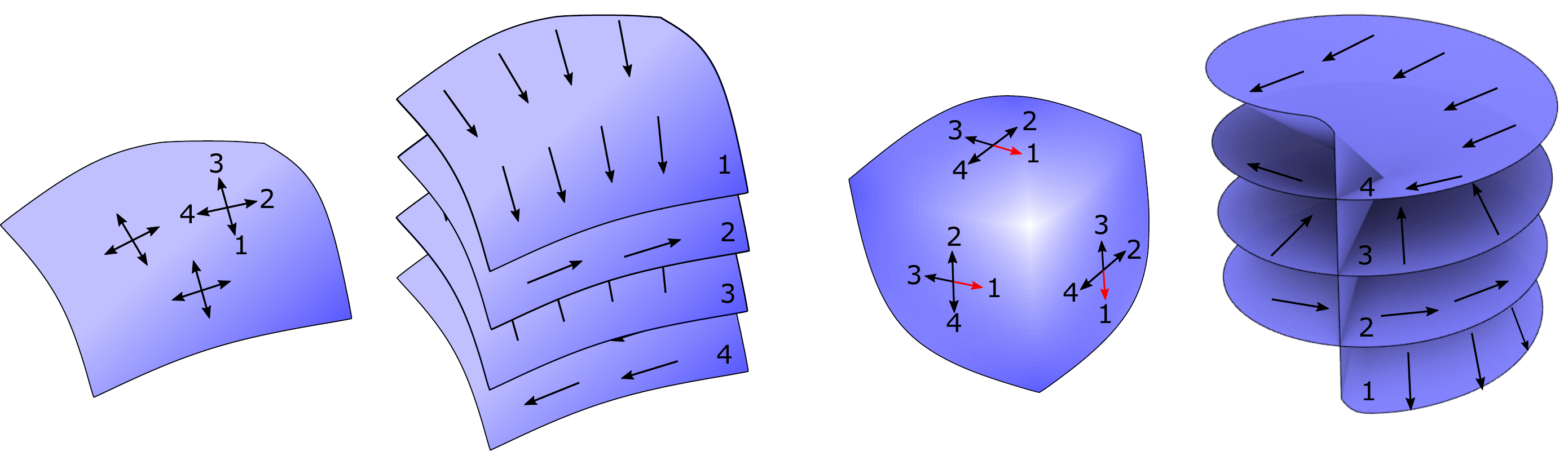}
		\put(9,-1){\small (a)}
		\put(35,-1){\small (b)}
		\put(61,-1){\small (c)}
		\put(87,-1){\small (d)}
	\end{overpic}
	\vspace{-4mm}
	\caption{4-direction frame fields and the corresponding cover spaces. (a)\&(b): a field without singular vertex and the four separate sheets of the cover space. (c)\&(d): a field with a singular vertex on the cube-corner shaped surface and the four sheets of cover space that are connected and coincide at the singular vertex.}
	\label{fig:cover_space}
	\vspace{-4mm}
\end{figure}

In addition, a vertex $x$ is singular if and only if it has a loop of neighboring vertices $[p_1,\cdots,p_n]$ such that $\mathbf{u}_{p_1}^i$ mapped by $\tau_{p_n,p_1}\circ\tau_{p_{n-1},p_{n}}\circ\cdots\circ\tau_{p_1,p_2}$ does not return to itself (Fig.~\ref{fig:cover_space}(c)).
Therefore on a patch containing no singular vertex, the transport $\tau_{x,y}$ remains the same regardless of the path taken between $x,y$ \cite{Crane:TrivialConnection} (Fig.~\ref{fig:parallel_frames}(b)).
On the other hand, the concentrated curvature at a singular vertex can only be multiples of $\frac{2\pi}{N}$, which explains the usage of $N$ symmetric directions: larger $N$ allows for more flexible singularities and flat connections.
We discuss the choice of $N$ later.

By solving for \textit{smooth} (or \textit{parallel}) frame fields that deviate minimally from the parallel transport and align to salient geometric features of the underlying surface \cite{Ray:GeometryAware,Crane:TrivialConnection,DirFieldEG2016} (see Appendix \ref{appn:smooth_field} for details), 
we obtain locally flat connections that closely approximate the linear connection while also having consistency among deformed shapes, therefore supporting improved feature learning by the extended surface convolutions.

While now we can translate tangent planes, another challenge unique to a surface domain rather than 2D images is the lack of canonical axes for the tangent planes. 
By randomly fixing an axis on one tangent plane, we risk significantly biasing the feature learning. Instead, a more robust approach is to sample several directions on the tangent planes as axes, and properly aggregate the learned features for the final output.
Fortunately, the $N$-direction frames provide a uniform sampling of tangent directions, which motivates  introducing their associated $N$-cover space that allows to organize the feature learning over multiple axes.

\vspace{-3mm}
\paragraph{$N$-cover space.}
A frame field induces an $N$-cover space over the domain surface \cite{quadcover,dubrovin2012modern}.
Intuitively, the cover space consists of $N$ copies $\mathcal{M}_i$, $i=1,\cdots,N$ of the base surface, with each copy $\mathcal{M}_i$ having a unit vector field $\mathbf{u}_x^{\sigma_x(i)}$, where $\sigma_x(i)$ indexes the vector of the $N$-direction frame at $x$ for the sheet $\mathcal{M}_i$; in addition, $\mathbf{u}_x^{\sigma_x(i)}$ and $\mathbf{u}_y^{\sigma_y(i)}$ are connected by $\tau_{x,y}$ the principal matching (Fig.~\ref{fig:cover_space}).
The unit vector field is well-defined everywhere on the cover space, except at singular vertices where different sheets of the cover space coincide.
In this paper, we use the vector field as the canonical axes and compute surface convolution on the cover space;
around singular vertices, our framework degenerates to a strategy similar to the parallel transport method \cite{Poulenard:2018:Multidirectional}.

\section{Surface based PFCNN}
\label{sec:model}

\subsection{Surface convolution via parallel frames}
\label{sec:pfconv}

Given a surface mesh $\mathcal{M}$ equipped with a parallel $N$-direction frame field $\mathbf{u}_{x}^i$, the surface convolution for a vertex $v_i$ with its feature vector $\mathbf{F}_{v_i}^j$, $j=1,\cdots,N$ on the $\sigma_{v_i}^{-1}(j)$-th cover sheet is computed by the following steps:
\begin{enumerate}
	\itemsep0em
	\item Choose $\mathbf{u}_{v_i}^j$ as the $x$-axis of the tangent plane $T_{v_i}\mathcal{M}$. Thus the local coordinate system is encoded by the $2\times3$ matrix $F_{v_i}^j = (\mathbf{u}_{v_i}^j, \mathbf{n}_{v_i}\times\mathbf{u}_{v_i}^j)^T$.
	\item For each vertex $v_k$ in the neighboring geodesic patch $\mathcal{N}_{v_i}$, project it onto the tangent plane as $\mathbf{v}_k'$ under coordinate system $F_{v_i}^j$. Let $\mathbf{u}_{v_k}^l = \tau_{v_i,v_k}(\mathbf{u}_{v_i}^j)$; the projected point has feature vector $\mathbf{F}_{v_k}^l$. Resample the projected feature map into a regular grid, denoted $\mathbf{F}_{\mathcal{N}_{v_i}}^j$.
	\item Convolve $\mathbf{F}_{\mathcal{N}_{v_i}}^j$ with regular kernels $\mathbf{K}$ defined under $F_{v_i}^j$. The responses constitute the feature vector of $v_i$ for the next network layer.
\end{enumerate}
Before presenting details for step 2, we remark that translation equivariance in the form of (\ref{eqn:euclidean_trans}) indeed holds locally:
\begin{equation}
\tau_{v_i,v_k}(f\star k) = \tau_{v_i,v_k}(f)\star k,
\end{equation}
where we assume $f$ is a function defined on the tangent plane $T_{v_i}\mathcal{M}$, and $k$ is the convolution kernel supported on tangent planes.
The equality holds because: on the left hand side, $f\star k$ returns a function on $T_{v_i}\mathcal{M}$ which is then transported by the flat connection $\tau_{v_i,v_k}$ to a function on $T_{v_k}\mathcal{M}$, while on the right hand side $f$ is first transported onto $T_{v_k}\mathcal{M}$ and then convolved with $k$ on $T_{v_k}\mathcal{M}$; since the transport $\tau_{v_i,v_k}$ only changes the underlying coordinate system bases, the functions $f,k$ when defined using local coordinates do not change at all by the transport, which makes the equality trivially true.
In addition, on a patch without singular vertices, the transport and equation holds regardless of the path taken between two vertices, which is different from the path-dependent parallel transport \cite{Poulenard:2018:Multidirectional};
for patches with singular vertices, because the transport minimizes deviation from the parallel transport (Sec.~\ref{sec:frame_field}), our convolution closely resembles the parallel transport approach.

\vspace{-3mm}
\paragraph{Projection to tangent space and resampling.}
Previous patch-based surface CNNs use various kinds of geodesic curve tracing to impose a polar coordinate system onto the neighborhood patch $\mathcal{N}_{v_i}$ and map each neighboring point onto the tangent plane \cite{GeoCNN:ICCV2015,AnisoCNN:NIPS2016,MoNet2017,Poulenard:2018:Multidirectional}. 
We follow a similar approach adapted from \cite{Budninskiy2018ParallelTU} that is simpler to compute and works even for point clouds, thus enabling easy extension of our framework to point clouds.
In particular, we modulate the geodesic coordinates computation using the local axes $F_{v_i}^j$, and re-triangulate the projected neighboring points with Delaunay triangulation to avoid flipped triangles, over which a regular grid in the shape of convolution kernels is then resampled and feature vectors interpolated.
The operation is encoded by a sparse tensor $S$ that is precomputed for a surface mesh and can be applied efficiently with standard NN libraries.
See Appendix \ref{appn:projection_resample} for details.

\subsection{PFCNN structures}
\label{sec:layers}

In this section we present the detailed structures of layers specific to PFCNNs. 
These layers can be combined with standard CNN layers and stacked into networks such as U-Net \cite{ronneberger2015u} and ResNet \cite{he2016deep}.

\textbf{Input layers.} 
The PFConv takes as input a group of $N$ feature maps corresponding to the $N$ cover sheets. These features can be constructed by simply duplicating the original input for $N$ copies, i.e. $|V|{\times} C_{in}\rightarrow |V|{\times} N{\times} C_{in}$, where $C_{in}$ is the input per-vertex feature length, or can be computed by further utilizing the local coordinate systems for different cover sheets. 
Indeed, we find that for tasks on deformable domains, e.g. non-rigid shape classification, segmentation and registration, a simple but effective input feature that is invariant to global rigid transformations is the normal vector and height from tangent plane in local coordinates, i.e. $\mathbf{F}_{v_k}^l {=} \left(F_{v_i}^j \mathbf{n}_{v_k}, \mathbf{n}_{v_i}^T\mathbf{n}_{v_k}, \mathbf{n}_{v_i}^T(\mathbf{v}_k - \mathbf{v}_i)\right)$ for each patch vertex $v_k \in \mathcal{N}_{v_i}$. 
In this case, the input layer constructs an expanded $|V|{\times} N{\times} H {\times} W {\times} C_{in}$ feature map directly by sampling the local features with regular grids (Sec.~\ref{sec:pfconv}), where $H{\times} W$ is the spatial shape of the subsequent convolution kernel to be applied.

\textbf{Output layers.}
For the final per-vertex output we need to reduce the grouped feature maps to an aggregation, i.e. $|V|{\times}N{\times}C \rightarrow |V|{\times}C$.
The reduction operation can be in different forms, e.g. taking the maximum or average among $N$ parallel channels, or being learned implicitly by a standard $1{\times}1$ convolution. 
The outputs can be further aggregated over all vertices into a single output for the whole shape, as in classification tasks.

\textbf{Convolution layers.}
Given an input feature map $\mathbf{F}_{in}$ of shape $|V|{\times} N{\times} C_{in}$, the convolution layer first
vectorizes it into $\textrm{vec}(\mathbf{F}_{in})$, multiplies with the sparse matrix $S$ of shape $(|V|{\times} N{\times} H {\times} W, |V|{\times} N)$ that does the feature map resampling (Sec.~\ref{sec:pfconv}), and reshapes the result vector into a tensor of shape $|V|{\times} N{\times} H {\times} W {\times} C_{in}$; it then multiplies with the convolution kernel of shape $H{\times} W{\times} C_{in}{\times} C_{out}$ to obtain the output feature map $|V|{\times} N{\times} C_{out}$.
In case the input layer provides an expanded feature map with local features as discussed above, the convolution is a simple multiplication with the kernel.
In addition, the special case of $1{\times}1$ convolution on each cover sheet through a $C_{in}{\times}C_{out}$ kernel skips the feature resampling step and is directly multiplied with $\mathbf{F}_{in}$ to obtain the output.
Note the same convolution kernel is shared for all $N$ cover sheets of feature maps, as the different cover sheets effectively sample the canonical axes over the surface domain.

\textbf{Pooling/Unpooling layers.}
Pooling and unpooling layers effectively change the spatial resolution of learned features.
For surface meshes the different domain resolutions can be built through a hierarchy of simplified meshes $\mathcal{M}_i$ with $\mathcal{M}_1 = \mathcal{M}$ and each coarse vertex $v \in V_{i+1}$ corresponding to a subset of dense vertices $\{v_k'\} \subset V_{i}$, using e.g. \cite{Garland:1997:QEM,Hoppe:1996:PM}.
We adapt the simplification process so that their $N$-direction frames are also mapped, i.e. $F_v^j$ corresponds to $F_{v_k'}^l$ the closest axes by rotation.
Pooling is then defined as $\mathbf{F}_{v}^j = \textrm{Pool}(\{\mathbf{F}_{v_k'}^l\})$, where $\textrm{Pool}(\cdot)$ takes channel-wise maximum or average; the layer has a signature of $|V_i|{\times}N{\times}C\rightarrow |V_{i+1}|{\times}N{\times}C$ in terms of feature map shapes. Unpooling is the inverse operation of pooling.

Throughout the paper we assume batch size one, although using larger batch size is trivial as long as each mesh of a batch has the same number of vertices on every domain resolution. 
We have implemented the above layers with Tensorflow; the code is publicly available\footnote{Code and data are available at https://github.com/msraig/pfcnn.}.

%% file: src/results.tex
\section{Experiments}
\label{sec:results}

We test the PFCNN framework and compare it mainly with the state-of-the-art MDGCNN \cite{Poulenard:2018:Multidirectional} on deformable domain tasks involving shape classification, segmentation and registration where MDGCNN achieves uniformly superior performances than other methods, and with the state-of-the-art TextureNet \cite{Huang_2019_CVPR} on the scene semantic segmentation task which has a rigid underlying domain.
We further do ablation study on the impact of parallel frames, cover space grouped feature maps and layer normalization, etc.

\subsection{Deformable domain tasks}
\label{sec:deformable_tasks}
For fair comparison, we use $5{\times}5$ convolution kernel for PFCNN and a larger 4(radial)$\times$8(angular) kernel for MDGCNN, and the same network structure for both methods in each task, except for registration where the same number of convolution layers are adopted. 
Network and training details are provided in Appendix \ref{appn:experiment_details}.

\vspace{-4mm}
\paragraph{Classification.}
The SHREC'15 non-rigid shape classification challenge \cite{shrec15} has 1200 shapes represented by surface meshes that belong to 50 categories.
We use a network with three levels of resolution and the localized normal vectors as input features (Sec.~\ref{sec:layers}) for PFCNN.
As shown in Table.~\ref{tab:shrec15_results}, our results outperform PointNet++~\cite{qi2017pointnetplusplus} even when it uses an ensemble of sophisticated input features, e.g. WKS and HKS, that are agnostic to non-rigid deformations.
We are on par with MDGCNN that uses as input the SHOT descriptor~\cite{tombari2010unique} which is rotation invariant and more sophisticated than our raw input.

\begin{table}
	\caption{Results on SHREC'15 non-rigid shape classification. PN+ is PointNet++\cite{qi2017pointnetplusplus}; ``raw'' means using spatial coordinates as input, ``en'' means using an ensemble of intrinsic shape descriptors. MDG is MDGCNN\cite{Poulenard:2018:Multidirectional} using SHOT features as input.}
	\label{tab:shrec15_results}
	\centering\footnotesize
	\begin{tabular}{|c|c|c|c|c|c|}
		\hline
		& PN+(raw) & PN+(en) & MDG & Ours  \\ \hline
		Accu.(\%) & 60.18 & 96.09 & 99.5 & 99.5  \\ \hline
	\end{tabular}
	\vspace{-1mm}
\end{table}

\vspace{-4mm}
\paragraph{Human body segmentation.}
The human body segmentation dataset proposed by \cite{Maron:2017:ToricCoverCNN} contains labeled meshes of diverse human identities and poses from various sources, split by 381/18 for training and testing. 
The meshes have very different scales which we normalize first. The mesh resolutions are also very different, with the number of vertices varying from 3k to 12k, yet our network works well on the these data without remeshing.
The network is a U-Net like structure with three levels of domain resolutions.

\begin{table}
	\caption{Results on human body segmentation. Our method outperforms MDGCNN on both original data and the remeshed data.}
	\label{tab:humanbody_segment_results}
	\centering\footnotesize
	\begin{tabular}{|c|c|c|c|c|}
	\hline
	& \multicolumn{2}{|c|}{Original} & \multicolumn{2}{|c|}{Remeshed} \\ \hline
	Method & MDGCNN & Ours & MDGCNN & Ours \\ \hline
	Accu.(\%) & 88.2 & 91.45 & 89.53 & 91.79  \\ \hline
	\end{tabular}
	\vspace{-3mm}
\end{table}

\begin{figure}[t]
	\begin{overpic}[width=\linewidth]{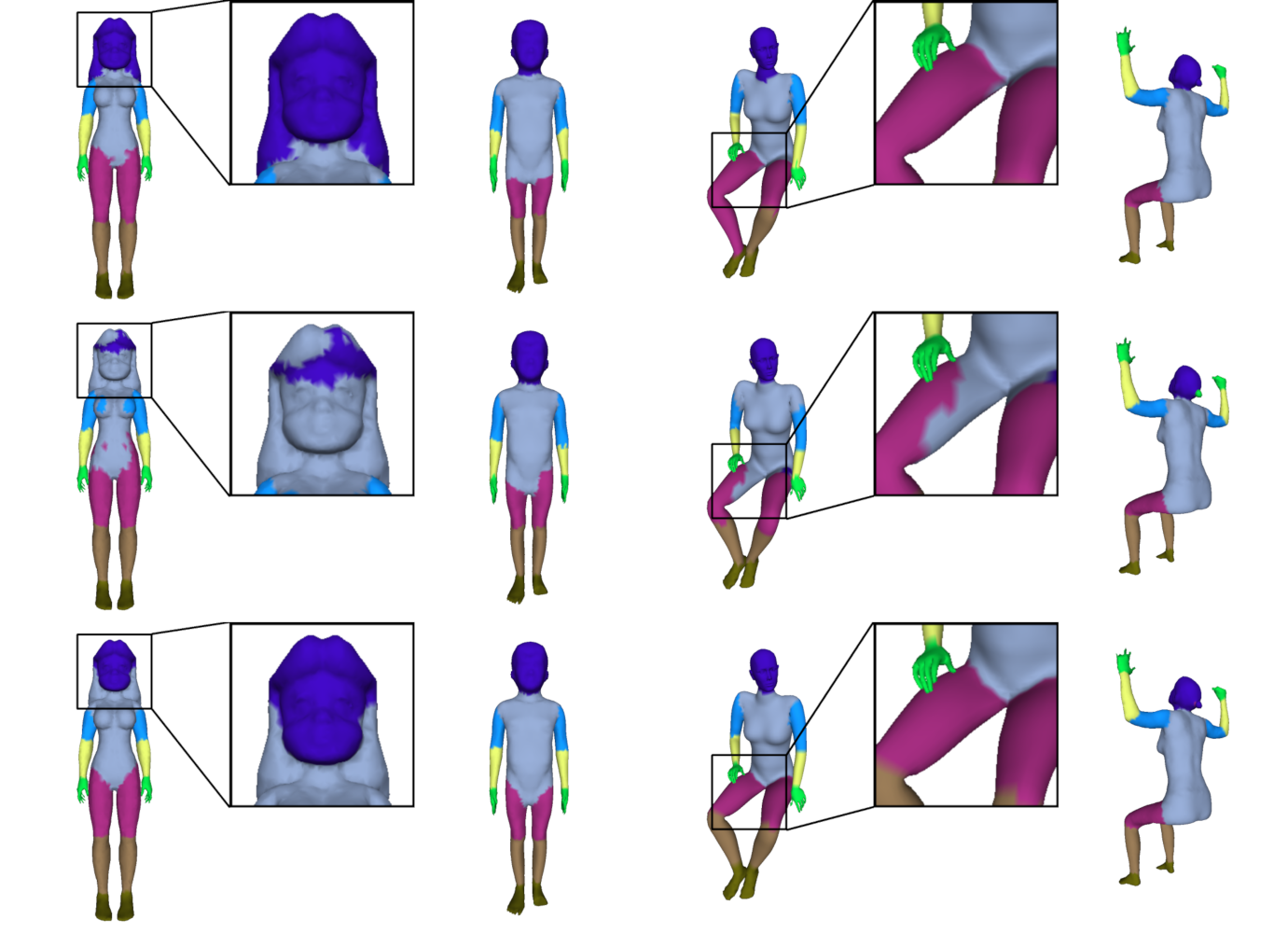}
		\put(-1,60){\small (i)}
		\put(-1,35){\small (ii)}
		\put(-1,13){\small (iii)}
	\end{overpic}
	\vspace{-6mm}
	\caption{Results of human body segmentation. (i) the ground-truth labeling; (ii) the results of MDGCNN; (iii) our results.}
	\label{fig:human_seg_results}
	\vspace{-5mm}
\end{figure}

To compare with MDGCNN, we test on both the original meshes and the resampled meshes generated with its open sourced code. 
Testing results are reported in Table~\ref{tab:humanbody_segment_results} and visualized in Fig.~\ref{fig:human_seg_results}. 
Note that the ground truth labeling for different samples are not always consistent, which hinders the possibility of achieving very high accuracy. For example, in Fig.~\ref{fig:human_seg_results} the third column GT mistakenly labels the shank to thigh. But our method correctly segments this part and has better coverage than MDGCNN. 
Still for some shapes which are dissimilar to the training data, e.g. the first column in Fig.~\ref{fig:human_seg_results} which has exceptional hair, both methods fail to segment the hair properly, although our method captures the face better.

\vspace{-4mm}
\paragraph{Human body registration by vertex classification.}
We test with the non-rigid human body registration task proposed by the FAUST dataset \cite{Bogo:CVPR:2014}.
In one scenario, the registration is achieved by classifying each input mesh vertex of a body shape into its corresponding vertex on the template mesh, as done in previous works \cite{GeoCNN:ICCV2015,AnisoCNN:NIPS2016,SplineCNN}.
We use a simple network consisting of a sequence of convolutions in the same level-of-detail for PFCNN, and a two-level network for MDGCNN, following their original setting.

The meshes in the FAUST dataset have the same topology with the template, which may be exploited unfairly to learn correspondences. 
Following MDGCNN, we remesh them to 5k vertices and different topologies. 
Using the nearest vertex as the correspondence between original meshes and the remeshed ones, we can get the ground truth vertex correspondence to the remeshed template, to supervise the registration task by classifying each vertex to 5k classes.

\begin{figure}[t]
	\begin{overpic}[width=\linewidth]{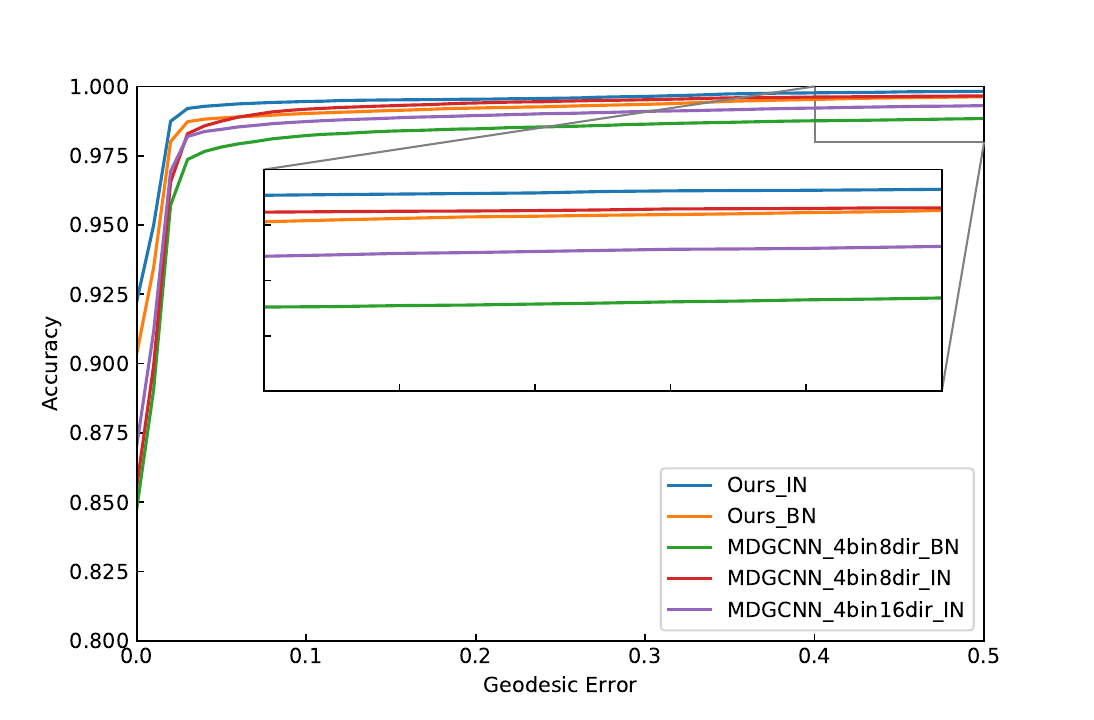}
	\end{overpic}
	\vspace{-6mm}
	\caption{Accuracy within given geodesic error for the non-rigid registration by vertex classification. }
	\label{fig:matching_accuracy}
	\vspace{0mm}
\end{figure}

We achieved $92.01\%$ accuracy on the remeshed data, as compared to $94.5\%$ accuracy on the original meshes.
To fully compare with MDGCNN, we also test variations of their networks with more radial bins and angular directions and different normalizations (more discussions in Sec.~\ref{sec:ablation}). 
The accuracies within bounded geodesic errors are plotted in Fig.~\ref{fig:matching_accuracy};
our results have even better zero-error accuracy than their best with 4$\times$16 kernels and instance normalization.
The visual results are shown in Fig.~\ref{fig:matching_results}; we can see that our results have a smoother mapping to the template shape.

In Appendix~\ref{appn:additional_comparison}, we test with a more challenging scenario of non-rigid registration by regression on noisy real scans with diverse and high genus meshes, where our results are again considerably better and more robust than MDGCNN.

To summarize, compared with the parallel transport based convolution by MDGCNN, using parallel frames that induce locally path-independent transport and the alignment to salient geometric features enables more efficient feature learning for our convolution; the difference is more obvious for finer scale tasks like segmentation and registration.

\begin{figure}[t]
	\begin{overpic}[width=\linewidth]{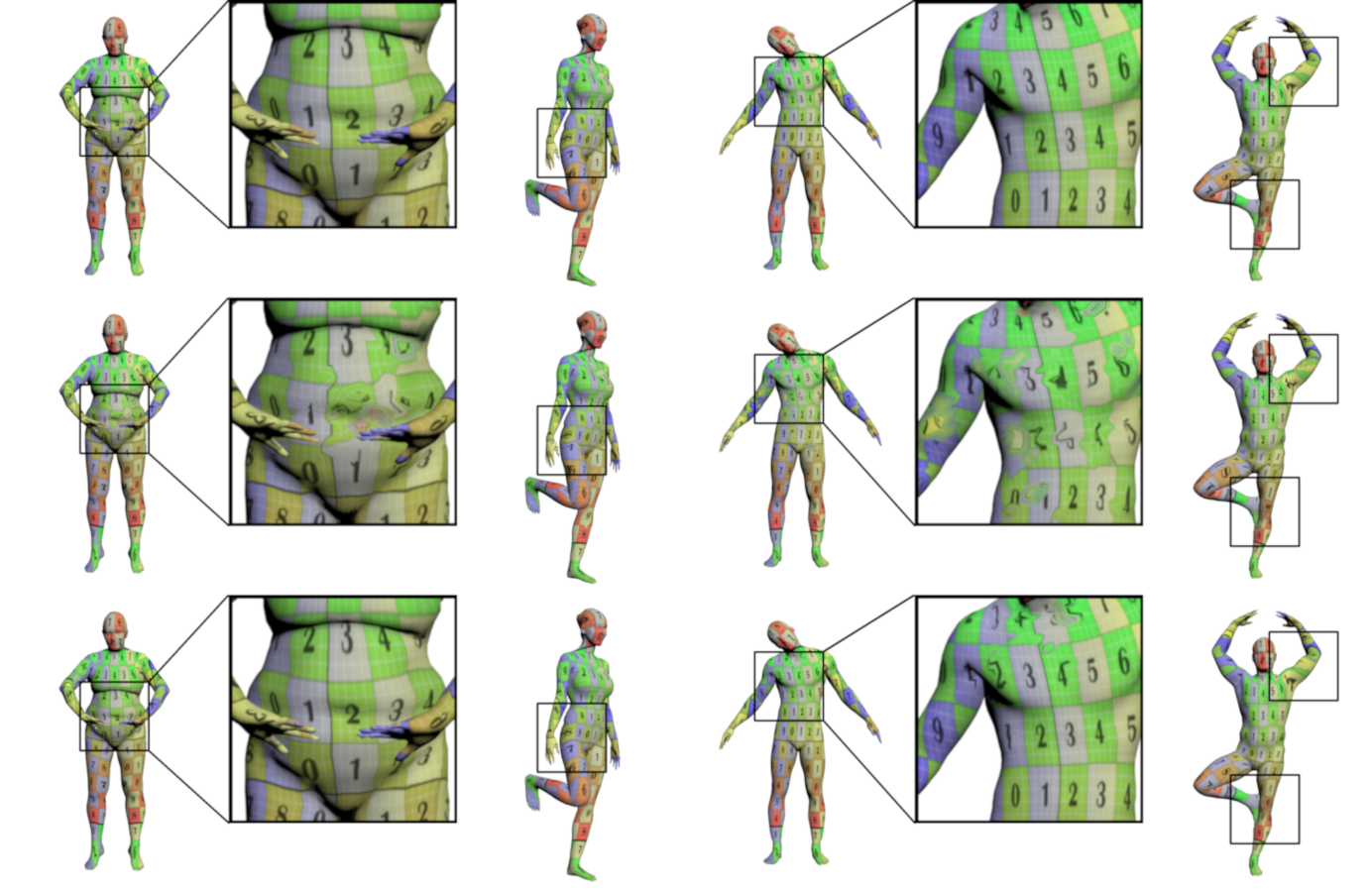}
		\put(-1,55){\small (i)}
		\put(-1,32){\small (ii)}
		\put(-1,10){\small (iii)}
	\end{overpic}
	\vspace{-6mm}
	\caption{Visual comparison of our method and MDGCNN on non-rigid registration. (i) the ground truth mapping; (ii) the best results of MDGCNN with 4 bins, 16 directions; (iii) our result.}
	\label{fig:matching_results}
	\vspace{-4mm}
\end{figure}

\subsection{Semantic scene segmentation}
\label{sec:scannet}

In this section, we evaluate on a widely used indoor scene semantic segmentation task provided by the ScanNet dataset \cite{dai2017scannet}.
While indoor scenes generally have rigid geometry dominated by flat walls and floors, PFCNN is still shown to achieve good performances, improving over the state-of-the-art TangentConv~\cite{Koltun:2018:TangentConv} and TextureNet~\cite{Huang_2019_CVPR} that use tangential and local patch convolutions.

We use a network with U-Net structure and three levels of domain resolutions.
We follow \cite{Huang_2019_CVPR} to prepare the training data by cropping small chunks from a whole scene and training on these chunks which are randomly rotated around the upright direction for augmentation. 
For network input, we follow \cite{Koltun:2018:TangentConv} to include the height above ground, normal vector, color and distance from the local tangent plane for each mesh vertex of a surface patch, rather than the localized normal vector as discussed in Sec.~\ref{sec:layers}, while \cite{Huang_2019_CVPR} uses additional high resolution texture images as input.
For fair comparison, we have used a network with similar amount of trainable parameters to \cite{Huang_2019_CVPR};
we also explore the effect of increasing the network size and report a better performance.

\begin{figure}[t]
	\includegraphics[width=\linewidth]{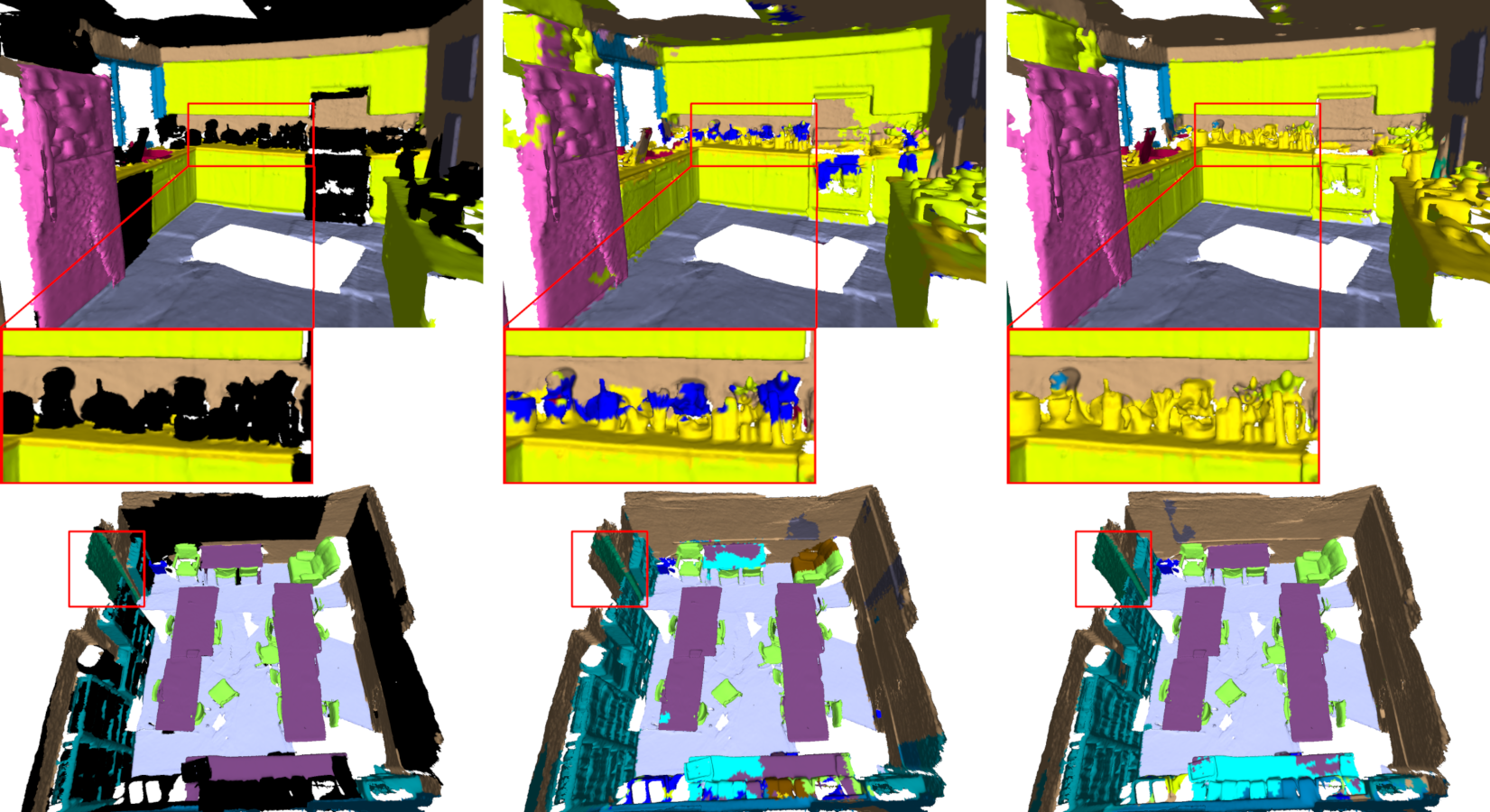}
	\vspace{-1mm}
	\put(32, 3){\small (i)}
	\put(113,3){\small (ii)}
	\put(195,3){\small (iii)}
	\caption{Example indoor scenes of ScanNet segmented by comparing methods. (i) is the ground truth segmentation; (ii) is the results of \cite{Huang_2019_CVPR}; (iii) shows our results. Our results have more regular boundaries separating regions of larger consistency.}
	\label{fig:scannet_visual_results}
	\vspace{-4mm}
\end{figure}

The result statistics of comparing methods and ours on validation sets are shown in Table~\ref{tab:scannet_segment_results}; our results have much better mean IoU and mean accuracy than theirs, which shows our network can better distinguish smaller objects than just the dominant segments like floors and walls.
Fig.~\ref{fig:scannet_visual_results} show some visual results. The black regions in (i) are unlabeled data; our method predicts reasonable labels for these regions. The boundaries separating different objects in our results are cleaner than \cite{Huang_2019_CVPR}, like the boundary between windows and the wall in the first row and the door and wall in the third row; our segments are also more regular and consistent.
See Appendix \ref{appn:additional_comparison} for more detailed data and visual results on both validation and test sets.

Considering that all three methods use tangent space convolutions, the results demonstrate that our locally translation equivariant convolution as the key difference is more effective in learning features.

\begin{table}
	\caption{Results on ScanNet segmentation task. mIoU is the class mean intersection over union. mA is the class mean accuracy. oA is the overall accuracy, which is significantly biased toward floors and walls that are dominant in scenes. Ours* uses a network with more convolution layers.}
	\label{tab:scannet_segment_results}
	\centering \footnotesize
	\begin{tabular}{|c|c|c|c|c|}
		\hline
		 & \cite{Koltun:2018:TangentConv} & \cite{Huang_2019_CVPR} & Ours & Ours* \\ \hline
		mIoU & 0.49 & 0.58 & 0.632 & 0.662 \\ \hline
		mA(\%) & 61.4 & 74.4 & 75.7 & 77.92 \\ \hline
		oA(\%) & 77.9 & 80.38 & 85.01 & 86.26 \\ \hline
	\end{tabular}
	\vspace{0mm}
\end{table}

\subsection{Ablation study}
\label{sec:ablation}

In this section, we evaluate how the core constructions and hyper parameters of PFCNNs affect performance.
We also study the impact of normalization on deformable domain tasks, as well as the behavior around singular vertices.

\begin{table}
	\caption{Testing accuracy of different convolution methods on the non-rigid registration task by vertex classification. PCF means using the principle curvature directions as tangent plane axes. PCF as FF means using the principal directions as the 4-direction frame field for our PFCNN framework.
	}
	\label{tab:comparing_conv}
	\footnotesize
	\centering
	\begin{tabular}{|c|c|c|c|}
		\hline
		&  PCF & PCF as FF & Ours \\ \hline
		Accu.(\%) & 83.29 & 89.80 & 92.01 \\ \hline
	\end{tabular}
	\vspace{-2mm}
\end{table}

\vspace{-4mm}
\paragraph{Using frames and grouped features.}
We evaluate the performances of different configurations that add components of the PFCNN construction one-by-one onto a baseline model. The evaluations are done on the task of human body registration by vertex classification (Sec.~\ref{sec:deformable_tasks}).
\vspace{-2mm}
\begin{itemize}
	\itemsep0em
	\item \textbf{Baseline model.} When using principal curvature frames as coordinate frames of the tangent plane, we have a baseline model similar to a bunch of recent previous works \cite{AnisoCNN:NIPS2016,MoNet2017,Zhong:DirectionalCNN:2017,Koltun:2018:TangentConv}.
	Using the network structure similar to PFCNN but without aligning the tangent planes by flat connections or feature map grouping by cover sheets, the trainable convolution kernel parameters are actually 16 times of PFCNN.
	However, the accuracy for this baseline configuration is 83.29$\%$ (Table~\ref{tab:comparing_conv}), much lower than PFCNN.
	\item \textbf{Principle curvature frames as 4-direction field.} As a modification to the baseline model, we regard the principal curvature frames as a 4-direction frame field and apply the PFCNN network. The result accuracy is 89.8$\%$ (Table~\ref{tab:comparing_conv}), much higher than the baseline model, while using only 1/16 trainable parameters.
	The improvement demonstrates that even if the frame field is not globally optimized to be smooth or aligned to salient features, by using its encoded flat connections that enable local translation equivariance and its induced cover space feature maps that sample tangent directions, the feature learning is significantly improved.
	\item \textbf{Full PFCNN model.} By additionally optimizing for a parallel frame field that aligns to geometric features, the PFCNN framework further improves to 92.01$\%$ registration accuracy (Table~\ref{tab:comparing_conv}).
\end{itemize}
\vspace{-3mm}

\begin{table}
	\caption{Accuracy and runtime cost of different frame field symmetry orders $N$, on the non-rigid registration task by vertex classification. The costs are measured on an RTX2080 GPU.}
	\label{tab:comparing_different_symorders}
	\centering\footnotesize
	\begin{tabular}{|c|c|c|c|c|c|}
		\hline
		$N$  & 1 & 2 & 4 & 6 & 8 \\ \hline
		Accuracy(\%) & 83.11 &91.81 &92.01 &92.45 &93.35  \\ \hline
		Time($ms$) & 56.81 &87.57 &139.10 &183.91 & 227.72 \\ \hline
		Memory(MB) & 148.67 & 156.15 & 205.81 & 371.45 & 409.18 \\ \hline
	\end{tabular}
	\vspace{-1mm}
\end{table}

\vspace{-3mm}
\paragraph{Frame symmetry order.}
As discussed in Sec.~\ref{sec:frame_field}, when the rotational symmetry order $N$ of the frame field gets larger, the frame field has more flexibility to achieve both smoothness and alignment to salient features. 
However, an increased $N$ also leads to larger computational cost, as the size of feature maps to compute increases too. 

We tested the different $N$ values again with the registration by vertex classification task, but modified the network structure to make sure each group of the feature map has the same size (i.e. 64), so that for different $N$ the amount of trainable convolution kernel parameters remains the same.
The performances for different $N$ are shown in Table~\ref{tab:comparing_different_symorders}.
We can see that the choice of $N=4$ strikes a balance between accuracy and computational overhead:
for $N<4$ the accuracy is notably lower due to the limited field smoothness, and for $N>4$ the computational cost is higher, with the extra runtime roughly in proportion to the number of axes sampled. We have used $N=4$ for all the other experiments in this paper.

\vspace{-3mm}
\paragraph{Normalization.}
It is well known that normalization can speed up the training procedure and make it more stable.
Here we study the impact of different normalizations on surface based CNNs more closely.
For the registration by classification task, we test our method and MDGCNN with batch normalization (BN) and instance normalization (IN). 
Note that since the batch size is one, the difference between BN and IN is that, the channel wise statistics of moving mean and average are used in testing stage for BN but not IN.
The result is shown in Fig.~\ref{fig:matching_accuracy}. 
We find that with IN both our method and MDGCNN achieve better performances.
We repeat the experiments on the shape classification task (Sec~\ref{sec:deformable_tasks}); the result is shown in Table~\ref{tab:ablation_normalization}. 
From all these experiments, we can see that IN is better than BN for these tasks on deformable domains. 
We argue that this is because the diversely deformed shapes do not share common statistics of channel-wise mean and variance, akin to the observation in image style transfer \cite{perez2018film,Dumoulin2016ALR,huang2017arbitrary} that these statistics encode styles rather than content.

\begin{table}
	\caption{Classification accuracy with different normalization.}
	\label{tab:ablation_normalization}
	\vspace{0mm}
	\footnotesize
	\centering
	\begin{tabular}{|c|c|c|c|c|c|c|}
		\hline
		 & \multicolumn{2}{|c|}{Ours} & \multicolumn{2}{|c|}{MDGCNN} \\ \hline
		Normalization & BN & IN & BN &  IN \\ \hline
		Accuracy(\%) & 11 & 99.5 & 14.0  & 99.5  \\ \hline
	\end{tabular}
	\vspace{-3mm}
\end{table}

\vspace{-3mm}
\paragraph{Frame field singularity.}

\begin{figure}[t]
	\centering
	\begin{overpic}[width=0.9\linewidth]{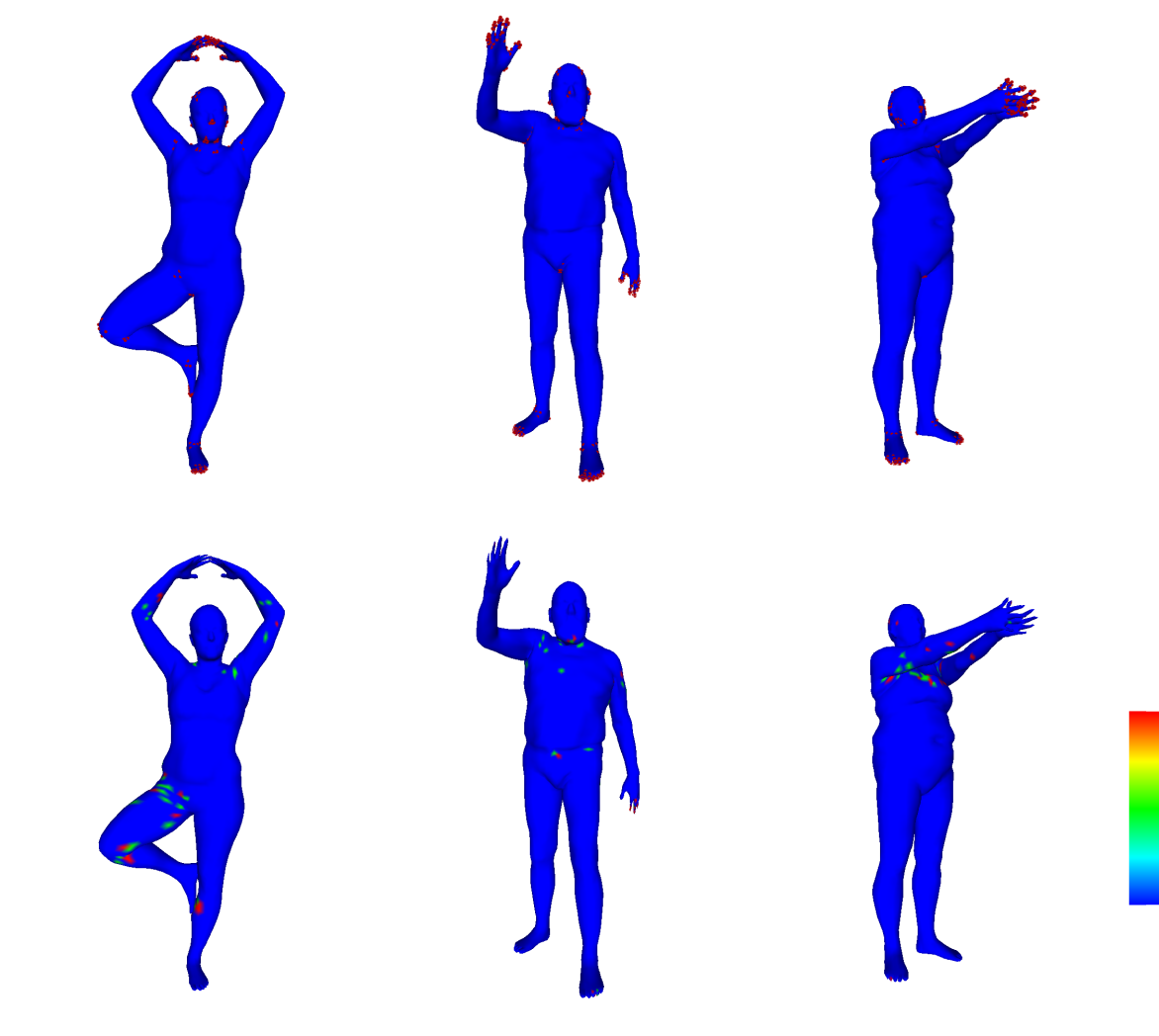}
		\put(0,70){\small (i)}
		\put(0,20){\small (ii)}
	\end{overpic}
	\vspace{-4mm}
	\caption{(i) shows the singular vertices in red and (ii) shows the prediction error map. There is no clear correlation between the singular vertices and the erroneous predictions.}
	\label{fig:singularity}
	\vspace{-5mm}
\end{figure}

As discussed in Sec.~\ref{sec:pfconv}, near singularities the translation equivariance is no longer path-independent, but our scheme degenerates to being similar to MDGCNN using parallel transport.
To find the relationship between the singularity of vertices and prediction error, we compare the distribution of singular vertices and the error map of geodesic distance between predicted vertex and the ground truth correspondence vertex. The distribution is shown in Fig.~\ref{fig:singularity}; we can see that the singular vertices mainly distribute on the nose, fingers or toes but the error maps of different shapes do not reflect these similarity.
We also compare the accuracy of singular vertices and all vertices in the registration by classification task. 
In particular, on the original dataset and the remeshed dataset, the registration accuracy of singular vertices versus that of all vertices are, $93.4\%/94.5\%$ and $90.2\%/92.2\%$, indicating no clear correlation of singular vertices and prediction errors.
Such robustness can be attributed to the degenerated convolution with path-dependent translation equivariance, the consistency of singularities across shapes (see Appendix \ref{appn:smooth_field}) and the capability of learned filters.

%% file: src/conclusion.tex
\section{Conclusion}

We have presented a surface mesh based PFCNN framework that closely mimics the standard image based CNNs and has local translation equivariance for convolutions.
It is enabled by using parallel $N$-direction frames that both encode flat connections on the surface to define path-independent translation, and sample tangent plane canonical axes to organize the convolutions by the $N$-cover spaces.
The PFCNNs are shown to be more effective at fine-scale feature learning than previous surface based CNNs.
In the future, we would like to investigate how the PFCNN framework can handle surface generation tasks, where the frame field also needs to be generated rather than precomputed.

%% file: src/appendix.tex
\appendix

\section{Overview}

In the appendix, we provide the detailed algorithms and additional discussions for computing the parallel $N$-direction frame fields, the mapping of neighborhood patches onto tangent planes and the resampling of feature maps there. 
The network structures and training details of experiments in the text, as well as additional results and comparisons with previous methods are also presented.

\section{Computing the parallel frame fields}
\label{appn:smooth_field}

Given a 3D surface mesh, the smooth or parallel frame field that approximates parallel transport of tangent spaces for neighboring points can be efficiently constructed \cite{DirFieldEG2016}.
In particular, we adopt the complex number based approach~\cite{Knoppel:2013:GOD,Diamanti:2014:PolyVector} to encode the $N$-direction fields.
We identify the tangent plane $T_xM$ with the complex plane, and a set of unit length vectors $\{u\cdot e^{ik\frac{2\pi}{N}}| k=0,\cdots,N-1\}\subset \mathbb{C}$ forming a rotationally symmetric $N$-direction frame can be conveniently encoded by their common $N$-th order power $z = u^N\in\mathbb{C}$. 
To compute a smooth frame field that a) deviates from the parallel transport minimally and b) aligns with salient geometric features of the domain surface, we solve the following optimization problem:
\begin{equation}
{\min_{\{z_i\}}}. \sum_{i\sim j}{\|z_i - t_{ji}z_j\|^2} + \lambda\sum_{i}{w_i\|z_i - z_i^0\|^2},
\end{equation}
where $i,j$ are neighboring vertices on the surface mesh, $t_{ji}\in\mathbb{C}$ is the discrete parallel transport along the edge $ij$ that rotates the tangent plane of $j$ to identify with that of $i$~\cite{Knoppel:2013:GOD}, $\lambda$ is the weight for the second curvature direction alignment term, $w_i = \tanh(|k_{max}-k_{min}|)$ measures the anisotropy at the $i$-th vertex using its maximum and minimum principle curvature values $k_{max},k_{min}$, and $z_i^0$ is the complex $N$-th order power of the maximum curvature direction at the vertex.
The first term is a discretization of the Dirichlet energy of the frame field that measures its variation and encourages parallelism.
The second term encourages alignment of the frame field to strong anisotropic directions and salient geometric features of the surface.

\begin{figure}[t]
	\includegraphics[width=\linewidth]{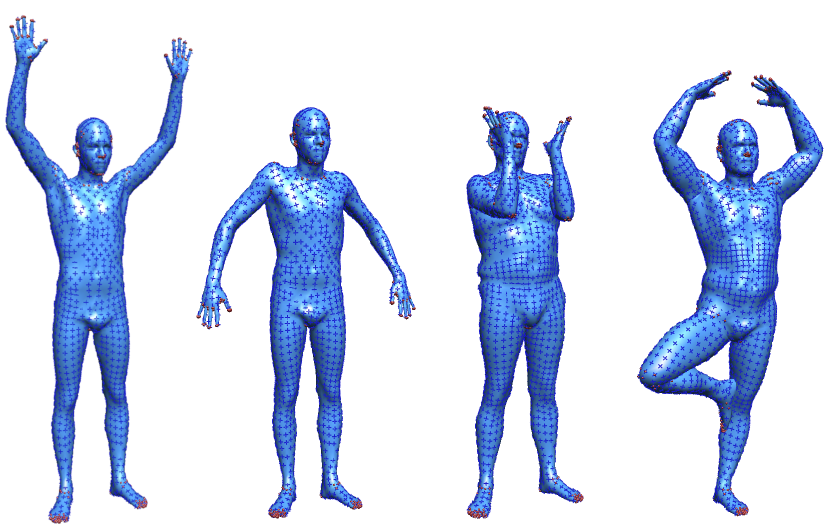}
	\vspace{-5mm}
	\caption{The smooth frame fields (shown as crosses) aligned with salient geometric features have strong consistency among diverse human body shapes. The singular vertices marked as red points are also distributed similarly across the shapes, concentrating on regions of high curvature, e.g. nose, finger tips, and toes.}
	\label{fig:smooth_field_examples}
	\vspace{-0mm}
\end{figure}

As shown in Fig.~\ref{fig:smooth_field_examples}, the smooth frame fields aligned with salient geometric features show strong consistency among deformed shapes, and the singular points are placed consistently at regions with high curvature.

\paragraph{Alignment to anisotropy.}
We test how different balances of field smoothness and alignment to strong anisotropy of the surfaces affect performances.
We generate four different sets of frames for the registration task, using $\lambda=0,0.01,0.1,1$ respectively. 
The testing accuracies are reported in Table~\ref{tab:smooth_align}, where ``SF'', meaning smooth frames without curvature direction alignment, corresponds to $\lambda=0$ and $\|z\|=1$ to prevent degenerate solutions.
From the results, we see that a mild alignment to strong surface anisotropic directions is helpful in achieving the best performances. Therefore, we have used $\lambda=0.01$ for all tasks shown in other parts of the paper.

\begin{table}
	\caption{Testing accuracy of different frame alignment choices, on the FAUST non-rigid registration task by classification. SF means smoothness only without alignment to surface anisotropy. The other numbers are used as the curvature direction alignment weight $\lambda$ for computing the smooth frame field.	}
	\label{tab:smooth_align}
	\vspace{1mm}
	\footnotesize
	\centering
	\begin{tabular}{|c|c|c|c|c|}
		\hline
		& SF  & 0.01 & 0.1 & 1  \\ \hline
		Accu.(\%) & 88.56 & 92.01 & 91.97 & 90.77  \\ \hline
	\end{tabular}
	\vspace{-2mm}
\end{table}

\section{Tangent plane projection and resampling}
\label{appn:projection_resample}

The algorithm for building the convolution structure on a local patch of a mesh vertex is illustrated in Alg.~\ref{alg:tangent_param} in pseudo code.
For each mesh vertex, the algorithm first does a flood searching of $K$ neighbor vertices in $O(K)$ and projects the vertices onto the tangent plane using local coordinate systems.
It then triangulates the projected vertices into a Delaunay triangulation in $O(K\log{K})$, 
and samples $H{\times}W$ grid points against the triangulation in $O(HW \log{K})$.
The sampled grid points are finally stored into the sparse tensor that will be reshaped as a sparse matrix and readily multiplied with feature maps in each convolution operation (Sec.~\ref{sec:layers}).
Note that all vertices can be processed in parallel.

In the algorithm we have abused notations slightly, using $t[0]$ of a tuple to represent the vertex, its index, and its spatial position; the exact meaning should be clear from context.
For a given level of domain resolution, the patch size parameter $d$ is set to be the average edge length of all meshes in the given level of the training dataset.

\begin{algorithm}
	\SetAlgoLined \KwIn{$v_i\in V$, frames $F$, transport $\tau$, patch side length $d$, conv kernel shape $H{\times}W$ } 
	\KwOut{updated sparse tensor $S$ of shape $|V|{\times}N{\times}H{\times}W{\times}|V|{\times}N$}
	
	\tcp{Flood to find and project neighbor vertices}
	$Q=[\left(v_i, (0,0), 0\right)], P=\{\}, visited=\{v_i\}$\;
	\While{$Q$ not empty}{
		$t = $dequeue($Q$), $P = P\cup t$\;
		\If{$\textrm{dist}(v_i, t[0])>\sqrt{2}d$ or $\|t[1]\| > \sqrt{2}d$}{
			continue\;
		}
		\For{$v_k \sim t[0], v_k\notin visited$}{
			$visited=visited\cup v_k$\;
			$\mathbf{u}_{v_k}^l = \tau_{t[0],v_k}(\mathbf{u}_{t[0]}^{t[2]})$\;
			$\mathbf{v}_k' = 0.5\cdot(F_{t[0]}^{t[2]}+F_{v_k}^l)(\mathbf{v}_k - t[0]) + t[1]$\;
			enqueue($Q$, $(v_k, \mathbf{v}_k', l)$)\;
		}
	}
	
	\tcp{Triangulate the projected points}
	$DT(P) =$ Delaunay triangulation of $\{t[1]|t\in P\}$\;
	
	\tcp{Resample with a regular grid sized $d{\times}d$}
	\For{$j=1,\cdots,N$}{
		\For{\normalfont{grid point} $p_{r,c}$, $1\leq r\leq H, 1\leq c \leq W$}{
			find the containing triangle in $DT(P)$ with vertices corresponding to $(t_a, t_b, t_c)\subset P$;
			compute barycentric weights $(w_a, w_b, w_c)$\;
			$S(i,j,c,r,t_a[0],t_a[2]) = w_a$\;
			$S(i,j,c,r,t_b[0],t_b[2]) = w_b$\;
			$S(i,j,c,r,t_c[0],t_c[2]) = w_c$\;
		}
	}

	\caption{Tangent plane projection and feature map resampling for a local patch} 
	\label{alg:tangent_param}
\end{algorithm}

\section{Network structures and training details}
\label{appn:experiment_details}

We have used convolution kernels with spatial size $5{\times}5$ for all deformable domain tasks, and $3{\times}3$ for the semantic scene segmentation.
All our networks have been trained with the Adam solver~\cite{AdamSolver} and batch size one, with fixed learning rate $10^{-4}$.

\begin{figure}[t]
	\includegraphics[width=\linewidth]{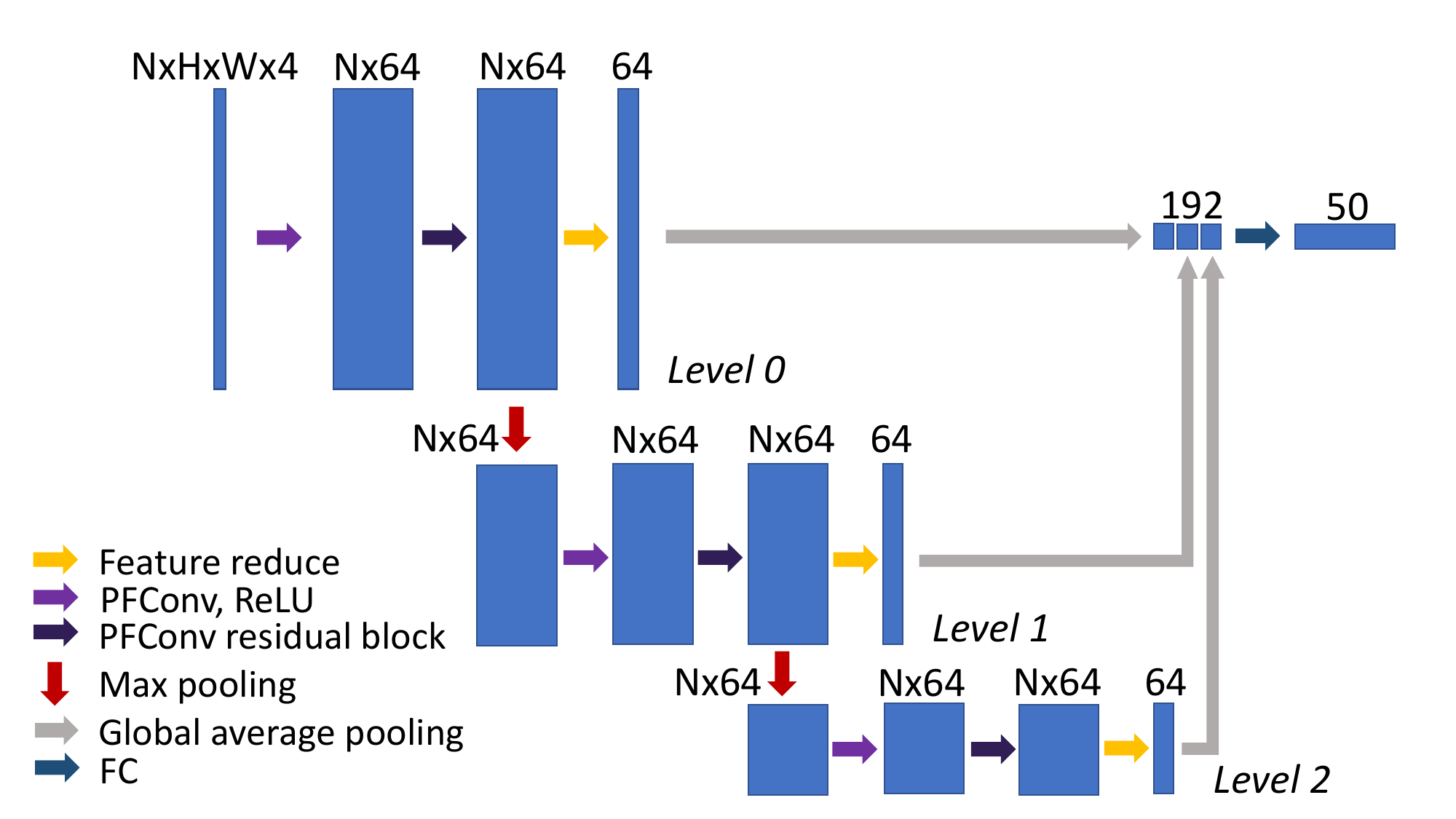}
	\vspace{-2mm}
	\caption{The network used for SHREC'15 non-rigid shape classification task. 
		Each box represents a feature map of shape $V{\times}C$, where $C$ is the total feature size for all $N{=}4$ cover sheets and given by numbers aside the boxes, and $V$ the number of surface vertices.
		The input feature map is a 4-channel feature of $H{\times}W$ grid points for each vertex (Sec.~\ref{sec:layers}).
		The ``convolution through residual block'' contains two sequential residual blocks, with each block made by two convolutions that retain the input feature size.
		All convolution operations except the last one are followed with instance normalization and ReLU. 
		Global average pooling is a standard average pooling over all vertices.	
		For this dataset there are around 10k, 1700, 300 vertices for the three level-of-details, respectively. 
	}
	\label{fig:shrec15_network}
	\vspace{0mm}
\end{figure}

The network structure used for SHREC'15 non-rigid shape classification is shown in Fig.~\ref{fig:shrec15_network}.
It is trained for 50 epochs on a single GPU. For the variant network without any normalization layers, it needs to train for 100 epochs until convergence. 

\begin{figure}[t]
	\includegraphics[width=\linewidth]{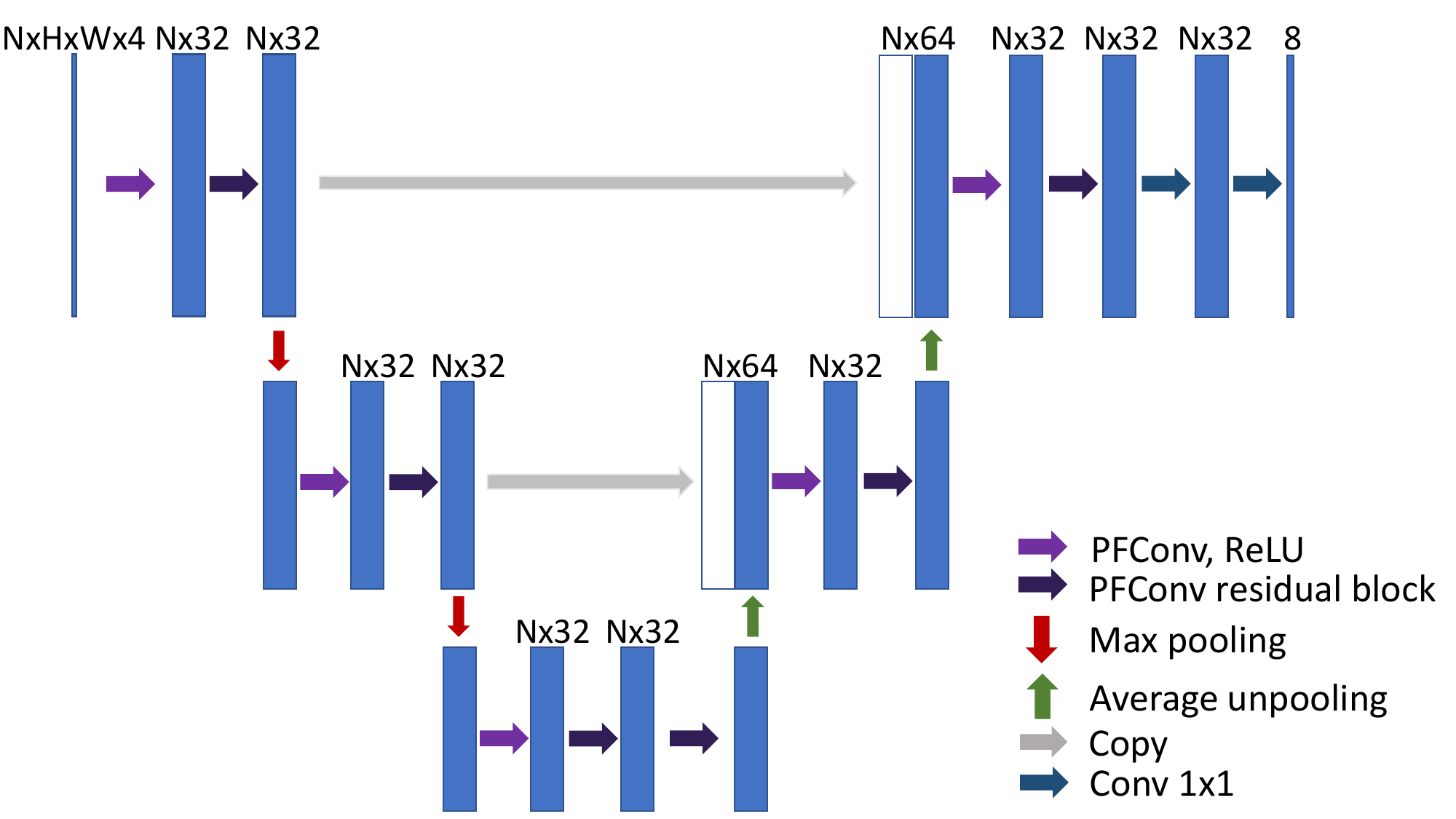}
	\vspace{-2mm}
	\caption{The network used for the human body segmentation task. See caption of Fig.~\ref{fig:shrec15_network} for detailed explanation. The number of vertices for the three level-of-details are $V, V/3, V/9$, where $V$ is the number of vertices of each mesh in the dataset. In the original dataset, $V$ varies from 3k to 12k. For the remeshed data, $V$ is around 7k.}
	\label{fig:human_seg_network}
	\vspace{-2mm}
\end{figure}

The network used for the human body segmentation task is shown in Fig~\ref{fig:human_seg_network}.
It has three level-of-details.
The loss function is the summation of cross entropy between predicted segmentation label and the ground truth label for each mesh vertex.
The network is trained for 50 epochs.
To obtain the predicted per-face segmentation labels, we sample points for each face of a test mesh and project the points onto closest vertices of our remeshed models, whose labels are used to vote for the face label of the original test mesh.

\begin{figure}[t]
	\center
	\includegraphics[width=0.8\linewidth]{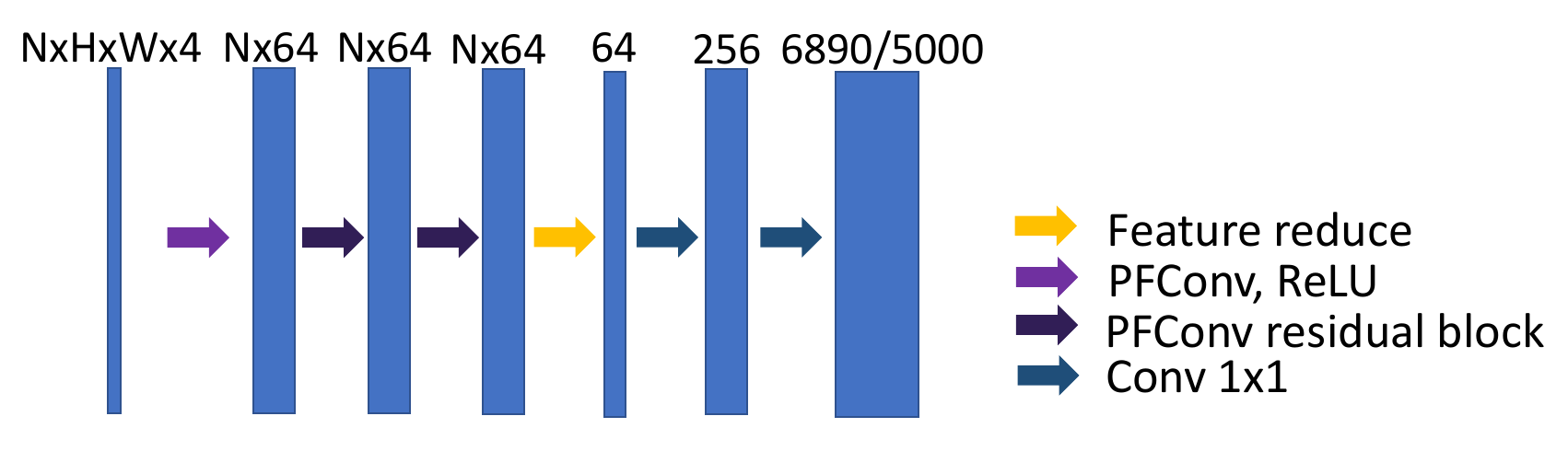}
	\vspace{0mm}
	\caption{The network used for human body registration through a classification of mesh vertices into 6890 or 5000. The number of surface vertices is 6890 for the original dataset and 5000 for the remeshed dataset. See caption of Fig.~\ref{fig:shrec15_network} for detailed explanation.
	}
	\label{fig:faust_classification_network}
	\vspace{-3mm}
\end{figure}

The network used for the human body registration task by vertex classification, 
and testing different frame field symmetry ordersis is shown in Fig.~\ref{fig:faust_classification_network}.
The network is trained for 400 epochs.

\begin{figure}
	\includegraphics[width=\linewidth]{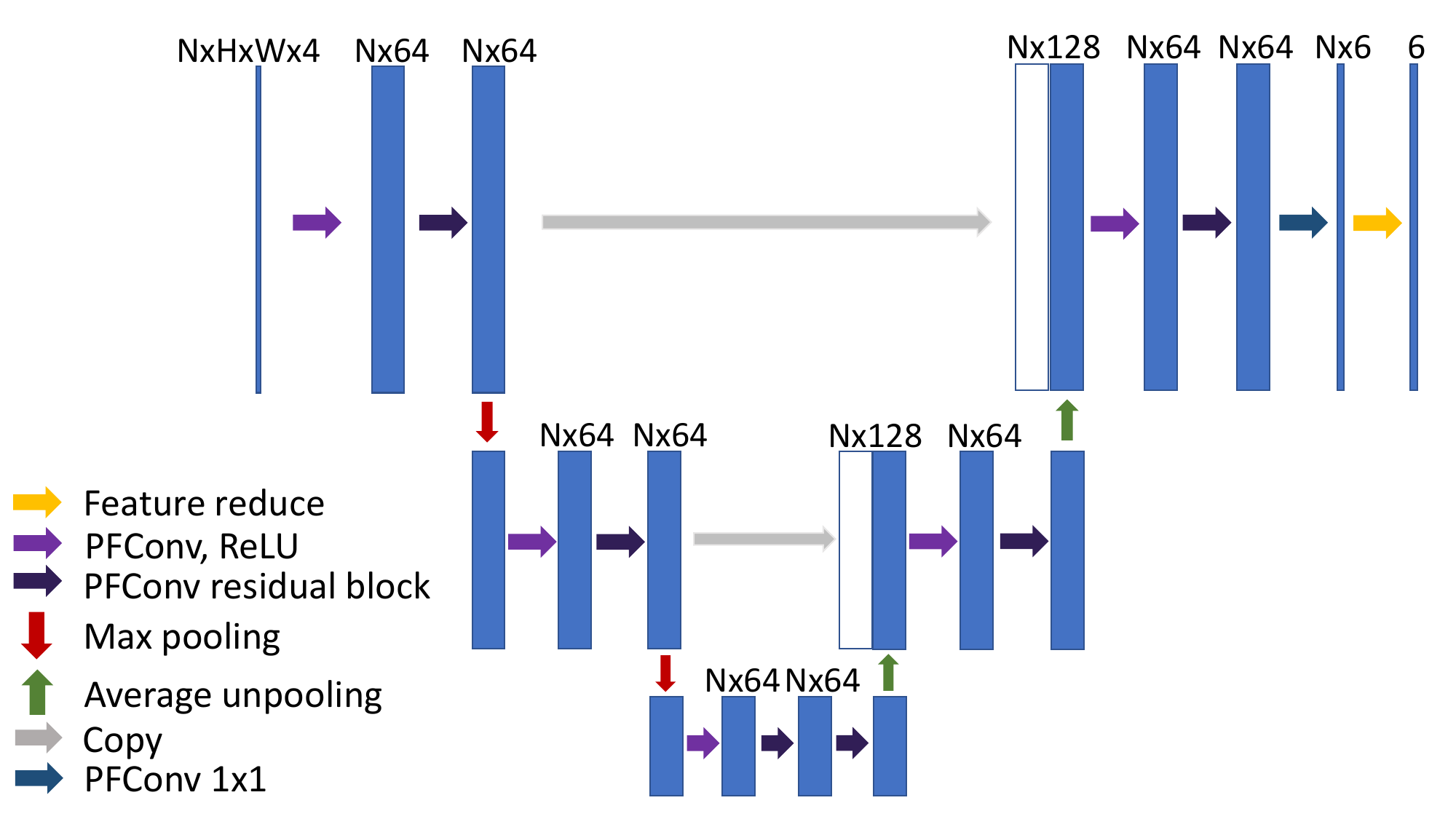}
	\vspace{-3mm}
	\caption{The regression network used for human body regression task. See caption of Fig.~\ref{fig:shrec15_network} for detailed explanation. The number of surface vertices are around 10k, 3.2k, 1k for the three level-of-details respectively.}
	\label{fig:faust_regression_network}
	\vspace{0mm}
\end{figure}

\begin{figure}[t]
	\includegraphics[width=\linewidth]{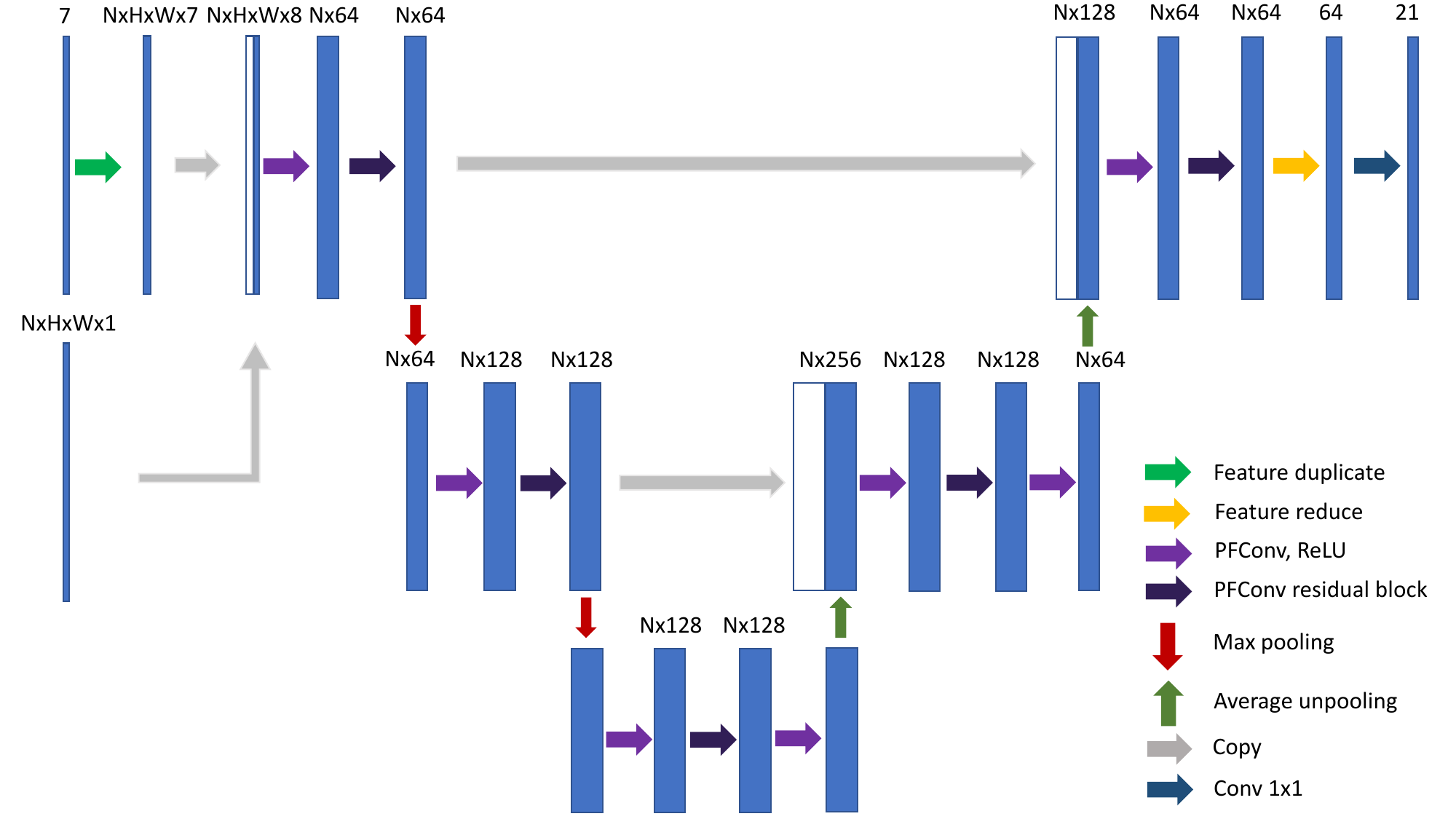}
	\vspace{-2mm}
	\caption{The network used for ScanNet segmentation task. See caption of Fig.~\ref{fig:shrec15_network} for detailed explanation. The input include the 7-channel feature of each vertex and the 1-channel local height feature for each grid point. The number of vertices for the three level-of-details are $V, V/3, V/9$, where $V$ is the number of vertices of each cropped chunk, with the crop method same as \cite{Huang_2019_CVPR}. }
	\label{fig:scannet_seg_network}
	\vspace{-2mm}
\end{figure}

The network used for the ScanNet semantic scene segmentation task is shown in Fig.~\ref{fig:scannet_seg_network}.
The network outputs, for each vertex, the probability distribution of 21 segmentation labels, which is compared with ground truth label using cross entropy during training. It is trained for 30 epochs.

\section{More results and comparisons}
\label{appn:additional_comparison}

\begin{figure}
	\centering
	\includegraphics[width=0.8\linewidth]{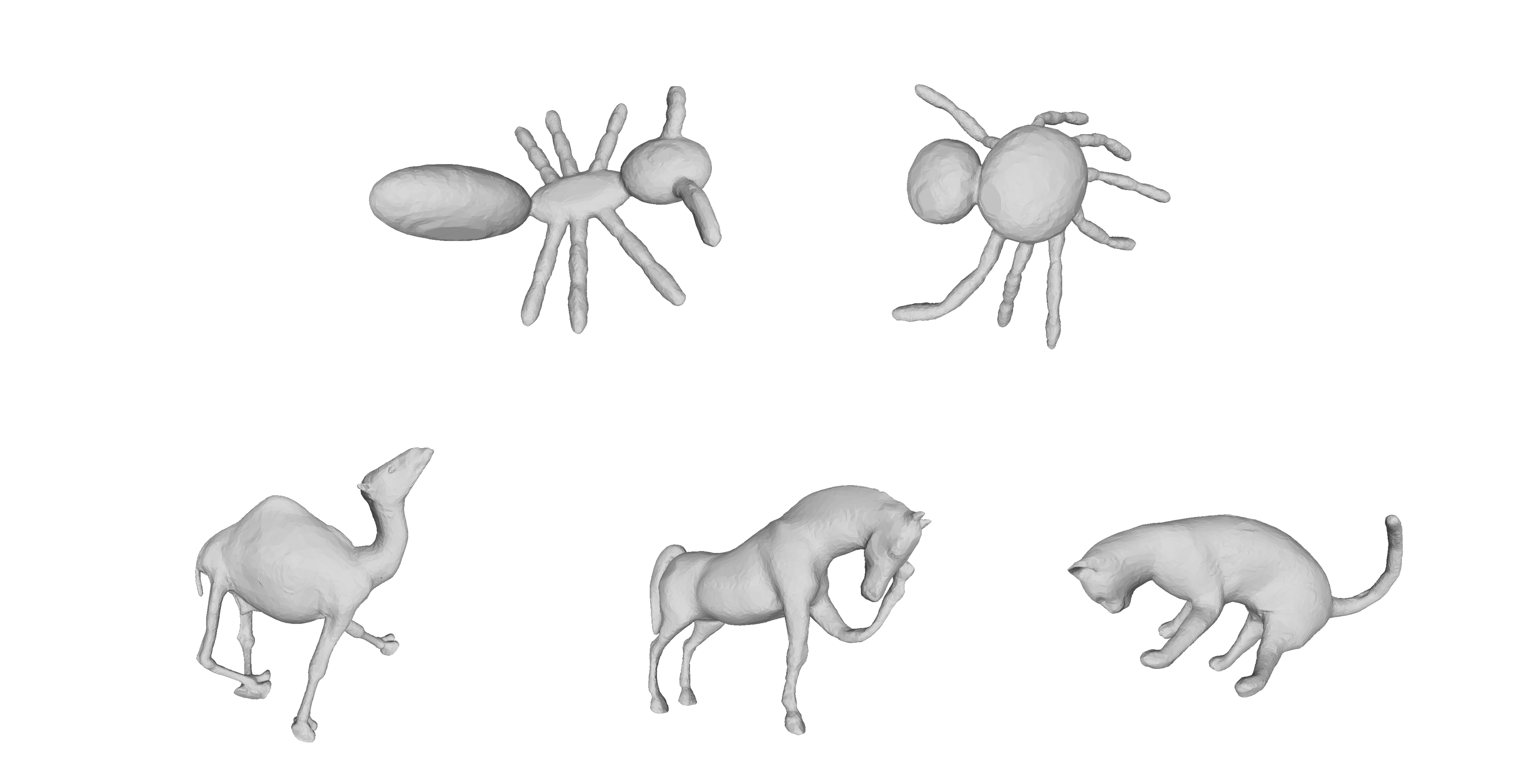}
	\vspace{-2mm}
	\caption{In the first row left shows the single incorrectly classified shape by our method; it is an ``ant'' misclassified as ``spider'' (an example shown on the right, is indeed confusingly similar to ``ant''). In the second row we show more shapes in the SHREC'15 dataset, which are ``camel'', ``horse'' and ``cat''. }
	\label{fig:classification_result}
	\vspace{-2mm}
\end{figure}

\paragraph{Shrec'15 classification}
We show some results in the classification task in Fig.~\ref{fig:classification_result}; the single misclassified shape by our method is a challenging ``ant'' that looks similar to the wrong label ``spider''.

\paragraph{Non-rigid registration by fitting template embedding.}
In this part we present an application that resolves the non-rigid registration problem with an approach different from the per-vertex classification (Sec.~\ref{sec:deformable_tasks}).
We notice that the registration by classification has severe limitations in real applications: to classify each vertex to 6k classes for example is not scalable when there are many input vertices, and the classification error does not measure at all how far away a mis-classified vertex is from ground-truth.
Thus we propose a novel but simple method for non-rigid registration that uses a surface-based CNN for direct regression of the template embedding in $\mathbf{R}^3$.

\begin{figure}[t]
	\centering
	\includegraphics[width=0.65\linewidth]{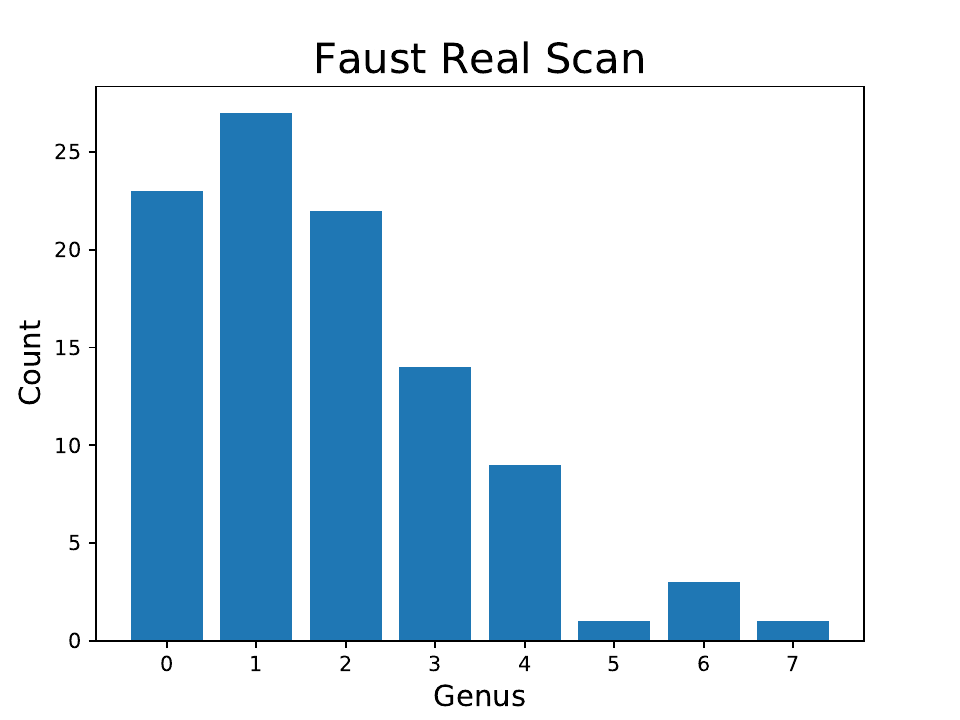}
	\caption{Genus of the meshes in the FAUST real scan dataset. More than half of the meshes have genus larger than 1.}
	\label{fig:faust_genus}
	\vspace{-2mm}
\end{figure}

\begin{figure}[t]
	\begin{overpic}[width=\linewidth]{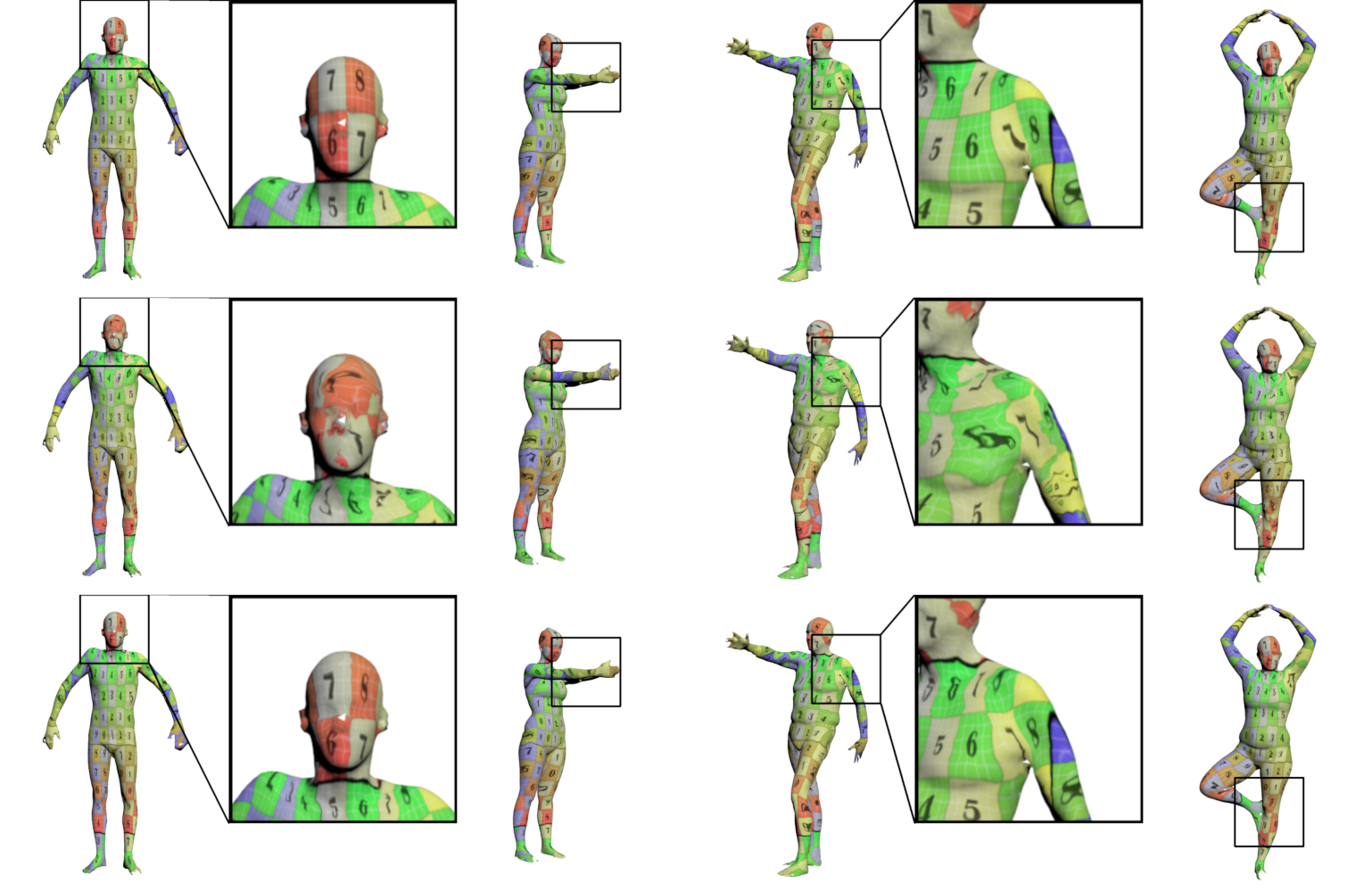}
	\put(-1,55){\small (i)}
	\put(-2,32){\small (ii)}
	\put(-3,10){\small (iii)}
	\end{overpic}
	\vspace{-4mm}
	\caption{Results of non-rigid human body registration through regression of the template embedding coordinates. (i) shows the texture mapping using the groundtruth correspondence. (ii) is the results of \cite{Poulenard:2018:Multidirectional}. (iii) shows our results.}
	\label{fig:faust_scanreg_results}
	\vspace{-2mm}
\end{figure}

\begin{figure}
	\centering
	\includegraphics[width=0.8\linewidth]{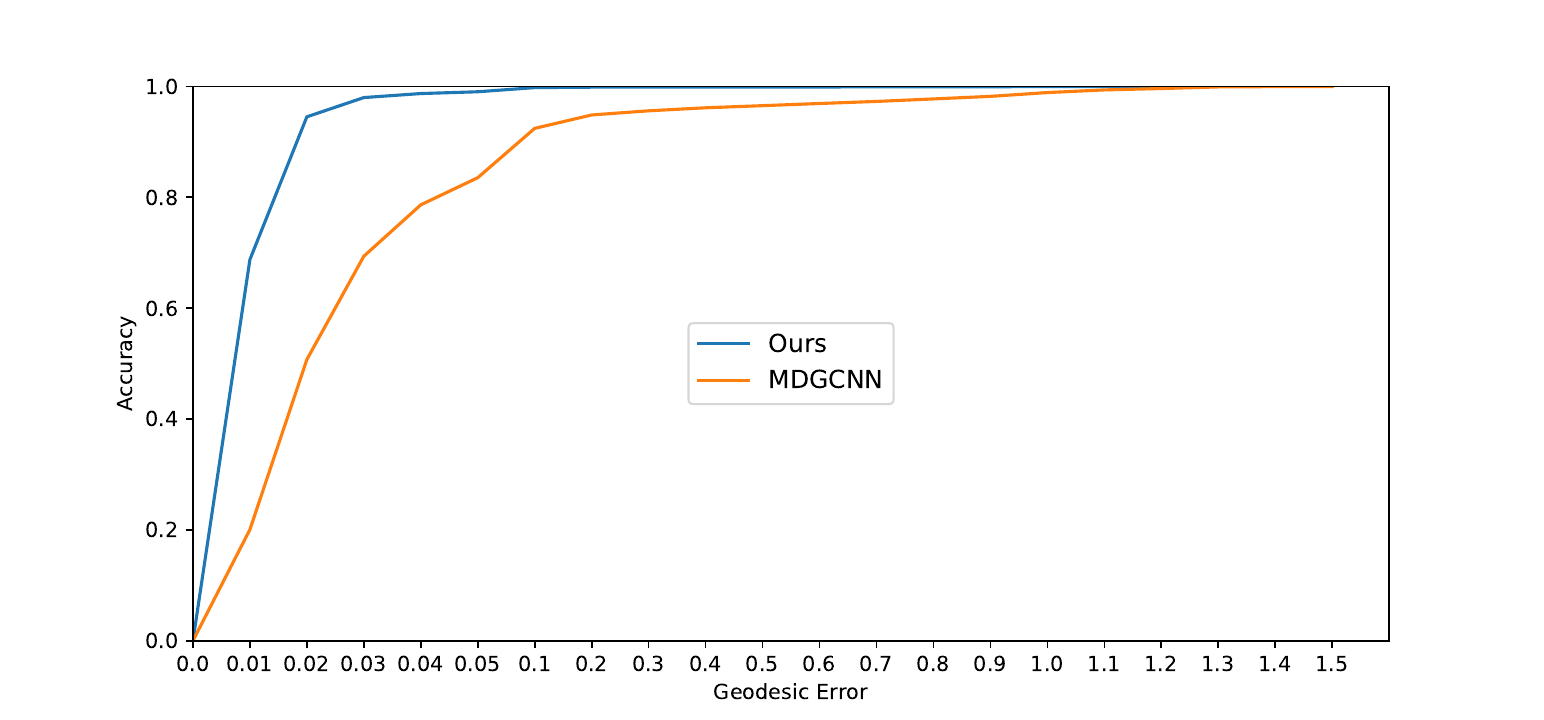}
	\vspace{-2mm}
	\caption{The ratio of vertices whose error is bellow given threshold. The per-vertex error is the geodesic distance between the predicted point position and ground truth on the template surface, normalized by square root of surface area. Our accuracy under 0.03 is $97.98\%$ while \cite{Poulenard:2018:Multidirectional} is $69.37\%$.}
	\label{fig:geodesic_error_distribution}
	\vspace{-2mm}
\end{figure}

We evaluate on the real scans of the FAUST dataset, which has 80 meshes for training and 20 for test.
The meshes are noisy, with diverse and high genus for different poses of a same person (see Fig.~\ref{fig:faust_genus} for statistics), which is frequently due to the merging of spatially intersecting components.
Since the raw scans are very dense meshes, we have remeshed each raw scan to simpler meshes with the number of vertices around 10k.
For each real scan there is a registered deformed template mesh, which provides the ground truth embedding for supervision and testing. 
To be specific, we project a vertex of the real scan to the closest point on the registered deformed template mesh, and take its position and normal vectors on the rest pose template as the supervising regression target.

The network for this point-wise regression is a standard UNet structure as shown in Fig.~\ref{fig:faust_regression_network}.
For each vertex of an input raw scan mesh, the output contains the position and normal vectors of the corresponding point on the rest pose template mesh.
The training loss is
\begin{align}
\vspace{-3mm}
L =& \frac{1}{V}\sum_{i}^V{\left(\|\mathbf{p}_i-\mathbf{p}_i^0\|_1 + \frac{w_{reg}}{V_i}\sum_{j\sim i}{\|\mathbf{p}_i-\mathbf{p}_j\|_1}\right)} \nonumber\\
&+ \frac{w_n}{V}\sum_i^V{\left(\|\mathbf{n}_i-\mathbf{n}_i^0\|_1 + \frac{w_{reg}}{V_i}\sum_{j\sim i}{\|\mathbf{n}_i-\mathbf{n}_j\|_1}\right)} \nonumber\\
&+ \frac{w_{con}}{E}\sum_{i\sim j}{|\mathbf{n}_i\cdot(\mathbf{p}_i-\mathbf{p}_j)|}, \nonumber
\vspace{-2mm}
\end{align}
where $V$ is the number of vertices of the raw scan mesh, $\mathbf{p}$ the regressed vertex position, $\mathbf{p}^0$ the target position, $\mathbf{n}$ the regressed vertex normal, $\mathbf{n}^0$ the target normal, $w_n = 0.1$ to normalize different scales between position and normal in the dataset, 
$w_{reg}=0.2$ the weight for Laplacian regularization terms of position and normal,
$w_{con}=20$ the weight for normal and position consistency,
$V_i$ the number of neighboring vertices of the $i$-th vertex,
and $E$ the number of directed mesh edges.
We use $l_1$ norm for these losses because there are noisy vertices in the raw scans which do not have valid target points on the template surface.
We train the network for 200 epochs on single GPU using Adam solver with a fixed learning $1\times10^{-4}$.

Geodesic errors of the network predictions on the test set are shown in Fig.~\ref{fig:geodesic_error_distribution}.
Following \cite{Kim:2011:BIM}, the geodesic error for a surface point $x$ with predicted position $y$ and ground truth point $y^*$ on the template surface $\mathcal{M}$ is computed as $\epsilon(x) = \frac{d_{\mathcal{M}}{(y,y^*)}}{\sqrt{|\mathcal{M}|}}$, where $d_{\mathcal{M}}{(\cdot,\cdot)}$ computes the geodesic distance of two points projected onto the surface $\mathcal{M}$, and $|\mathcal{M}|$ is its area for normalization.
Visual results are shown in Fig.~\ref{fig:faust_scanreg_results}. 
It is clear that our results are better than MDGCNN both quantitatively and qualitatively on these real scans, and the difference seems to be more obvious than the registration by vertex classification task on the clean meshes (Sec.~\ref{sec:deformable_tasks}).

\paragraph{More results of ScanNet segmentation.}

\begin{figure*}[t]
	\includegraphics[width=\linewidth]{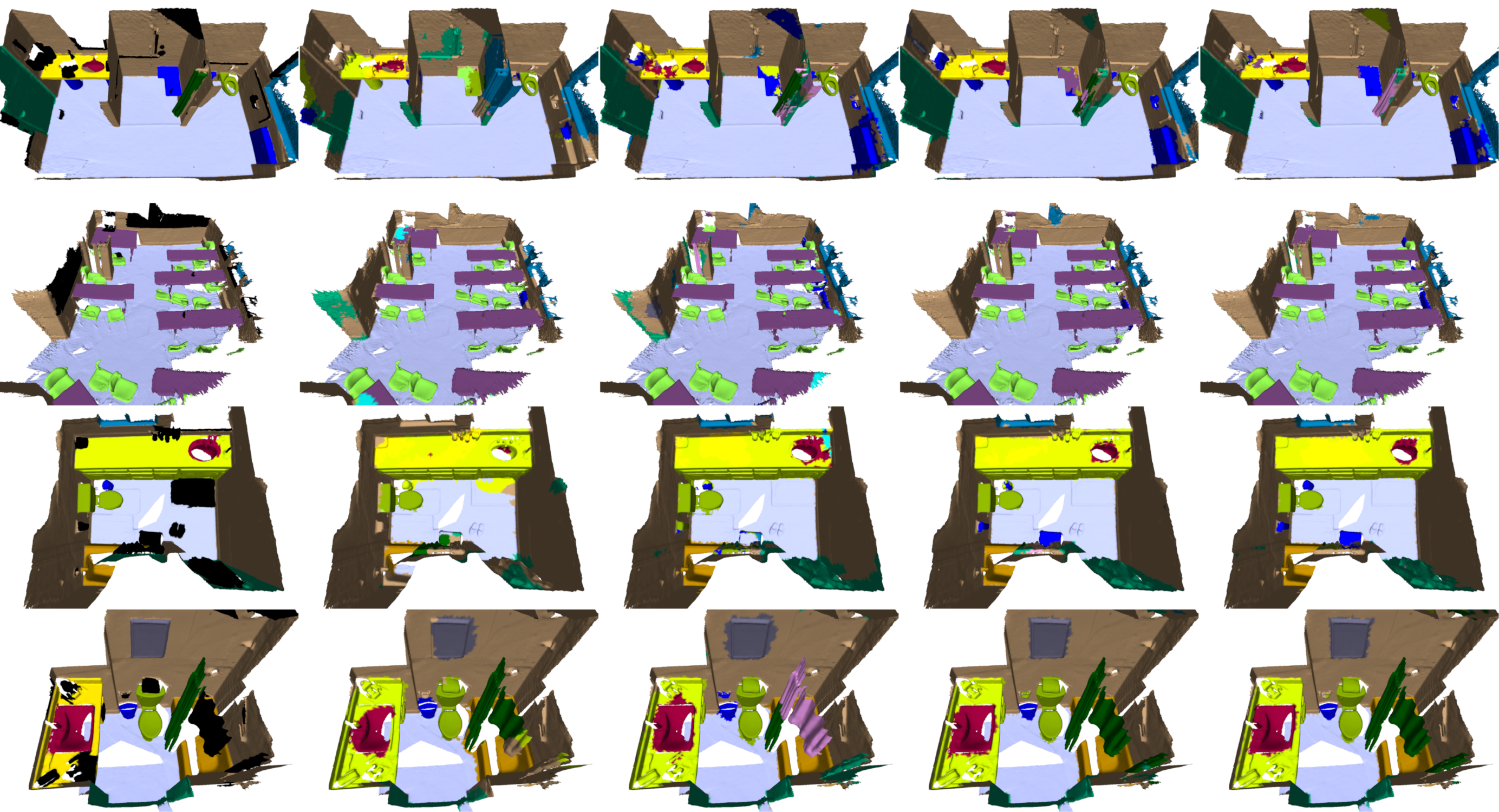}
	\vspace{-1mm}
	\put(50, 3){\small (i)}
	\put(150,3){\small (ii)}
	\put(250,3){\small (iii)}
	\put(350,3){\small (iv)}
	\put(450,3){\small (v)}
	\caption{More results of Scannet segmentation.(i) is the ground truth segmentation; (ii) is the results of \cite{Koltun:2018:TangentConv} (iii) is the results of \cite{Huang_2019_CVPR}; (iv) shows our results. (v) is the result of our method with deeper network. Our method gives clearer boundaries, like the boundary between window and wall in the second row, the boundary between picture and wall and the boundary of sink in the last row.}
	\label{fig:scannet_more_results}
\end{figure*}
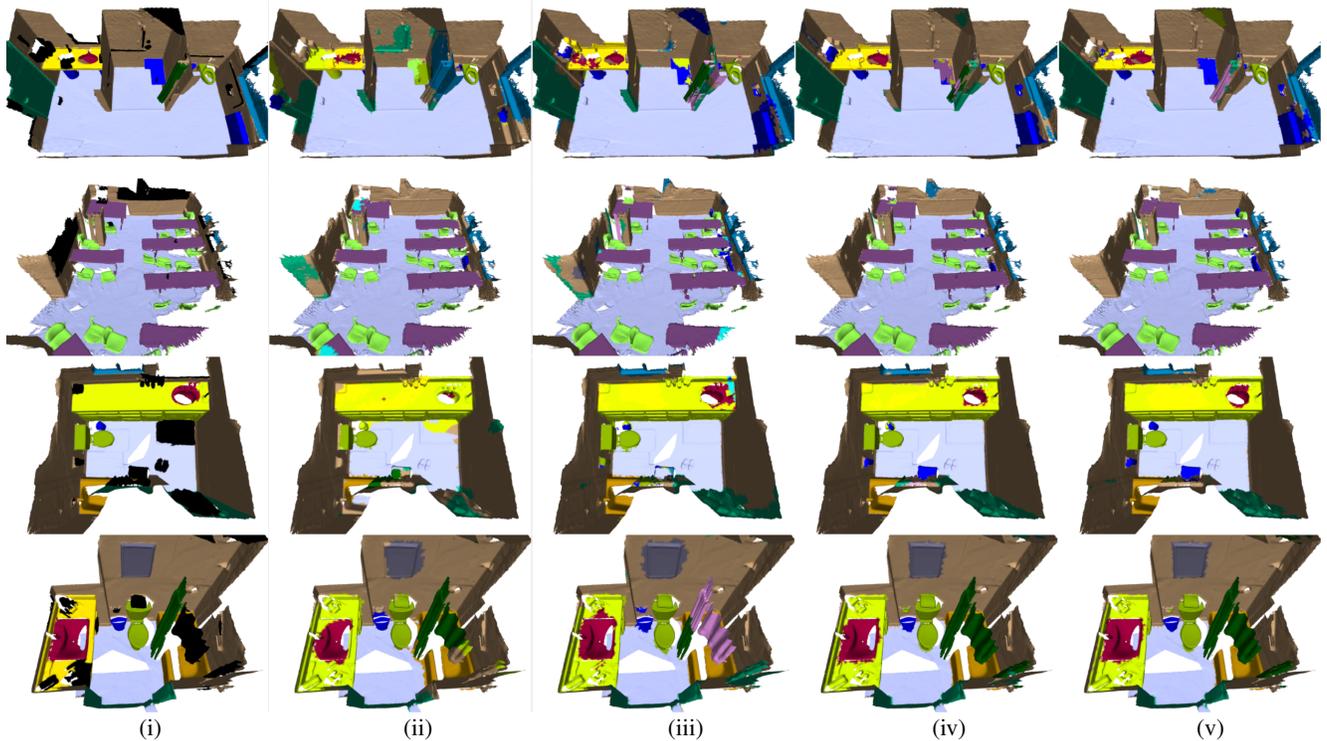

We present the per-category prediction accuracy (measured by IoU) of comparing methods for ScanNet semantic segmentation in Table~\ref{tab:scannet_val_category_results} and Table~\ref{tab:scannet_test_category_results}. For Ours*, the network structure is similar to the network shown in Fig.~\ref{fig:scannet_seg_network}, but each ``PFConv residual block'' contains three sequential residual blocks and the feature sizes in three levels are changed to 128, 256, 512, respectively.
More visual results are shown in Fig.~\ref{fig:scannet_more_results}.

\begin{table*}
	\centering
	\caption{Per-category IoU on ScanNet validation set. The abbreviations respectively stand for ``bathtub, bed, bookshelf, cabinet, chair, counter, curtain, desk, door, floor, otherfurniture, picture, refrigerator, shower, curtain, sink, sofa, table, toilet, wall, window''. The highest accuracies both among the three comparing results and among the four comparing results with our additional increased network are marked in bold. }
	\label{tab:scannet_val_category_results}
	 \footnotesize
	 \setlength\tabcolsep{3pt}
	\begin{tabular}{|c|c|c|c|c|c|c|c|c|c|c|c|c|c|c|c|c|c|c|c|c|c|}
		\hline
		Method & mIoU & bath & bed & book & cab & chr & cntr & crtn & desk & door & flr & other & pic & refrg & shwr & sink & sofa & tab & toil & wall & wdw \\ \hline
		\cite{Koltun:2018:TangentConv} & 49.1 & 68.0 & 63.8 & 56.3 & 41.7 & 73.6 & 45.6 & 33.2 & 40.7 & 34.9 & 91.9 & 26.2 & 14.5 & 31.7 & 28.1 & 44.2 & 62.8 & 51.5 & 68.8 & 67.9 & 38.3 \\ \hline
		\cite{Huang_2019_CVPR}& 58.1 & 67.6 & 67.3 & 71.3 & 46.8 & 78.1 & 44.4 & 52.5 & 47.5 & 44.8 & \textbf{94.4} & 40.2 & 21.1 & 35.2 & 51.3 & 51.7 & 64.0 & \textbf{63.5} & 80.3 & 75.0 & 46.0 \\ \hline
		Ours & \textbf{63.3} & \textbf{79.7} & \textbf{70.3} & \textbf{73.7} & \textbf{55.6} & \textbf{81.0} & \textbf{53.9} & \textbf{70.1} & \textbf{53.1} & \textbf{50.0} & 93.7 & \textbf{42.3} & \textbf{30.3} & \textbf{46.3} & \textbf{55.6} & \textbf{60.1} & \textbf{66.4} & 60.9 & \textbf{87.3} & \textbf{78.7} & \textbf{56.5} \\ \hline
		Ours* & \textbf{66.2} & \textbf{81.6} & \textbf{73.0} & \textbf{77.0} & \textbf{56.8} & \textbf{83.3} & \textbf{62.8} & \textbf{70.9} & \textbf{55.8} & \textbf{52.3} & 94.0 & \textbf{46.4} & \textbf{33.1} & \textbf{51.7} & \textbf{60.9} & \textbf{61.2} & \textbf{72.3} & \textbf{65.0} & \textbf{87.7} & \textbf{80.0} & \textbf{58.6} \\ \hline
	\end{tabular}
	\vspace{0mm}
\end{table*}

\begin{table*}
	\caption{Per-category IoU on ScanNet test set. See caption of Table~\ref{tab:scannet_val_category_results} for explanations.}
	\label{tab:scannet_test_category_results}
	\centering \footnotesize
	\setlength\tabcolsep{3pt}
	\begin{tabular}{|c|c|c|c|c|c|c|c|c|c|c|c|c|c|c|c|c|c|c|c|c|c|}
		\hline
		Method & mIoU & bath & bed & book & cab & chr & cntr & crtn & desk & door & flr & other & pic & refrg & shwr & sink & sofa & tab & toil & wall & wdw \\ \hline
		\cite{Koltun:2018:TangentConv} & 43.8 & 43.7 & 64.6 & 47.4 & 36.9 & 64.5 & 35.3 & 25.8 & 28.2 & 27.9 & 91.8 & 29.8 & 14.7 & 28.3 & 29.4 & 48.7 & 56.2 & 42.7 & 61.9 & 63.3 & 35.2 \\ \hline
		\cite{Huang_2019_CVPR} & 56.6 & 67.2 & 66.4 & 67.1 & 49.4 & 71.9 & \textbf{44.5} & 67.8 & 41.1 & 39.6 & 93.5 & 35.6 & 22.5 & 41.2 & \textbf{53.5} & 56.5 & 63.6 & 46.4 & 79.4 & 68.0 & 56.8 \\ \hline
		Ours & \textbf{60.2} & \textbf{74.6} & \textbf{71.2} & \textbf{67.4} & \textbf{53.5} & \textbf{75.6} & 41.6 & \textbf{68.1} & \textbf{42.0} & \textbf{43.4} & \textbf{93.8} & \textbf{40.1} & \textbf{27.0} & \textbf{51.2} & 51.1 & \textbf{61.2} & \textbf{69.4} & \textbf{48.3} & \textbf{84.7} & \textbf{77.7} & \textbf{61.5} \\ \hline
		Ours* & \textbf{62.2} & \textbf{79.7} & 69.7 & \textbf{75.0} & \textbf{57.7} & \textbf{79.2} & \textbf{47.6} & \textbf{68.5} & 36.6 & \textbf{46.8} & \textbf{94.2} & \textbf{41.4} & \textbf{30.7} & \textbf{53.2} & 49.4 & \textbf{68.1} & \textbf{71.5} & 47.5 & \textbf{88.0} & \textbf{79.6} & 59.3 \\ \hline
	\end{tabular}
	\vspace{0mm}
\end{table*}